\documentclass{article} 
\usepackage{iclr2027_conference,times}

\usepackage{amsmath,amsfonts,bm}

\def\eqref#1{equation~\ref{#1}}

\def\1{\bm{1}}

\DeclareMathAlphabet{\mathsfit}{\encodingdefault}{\sfdefault}{m}{sl}
\SetMathAlphabet{\mathsfit}{bold}{\encodingdefault}{\sfdefault}{bx}{n}

\usepackage{booktabs}
\usepackage{graphicx}
\usepackage{multirow}
\usepackage{subcaption}
\usepackage[table]{xcolor}
\definecolor{loopgreen}{HTML}{B2E0B3}
\definecolor{gmgreen}{HTML}{0B7A34}

\definecolor{loopyellow}{HTML}{F0E7A8}
\definecolor{loopred}{HTML}{F8C9C1}
\usepackage{amssymb}
\usepackage{wrapfig}
\newsavebox{\teaserboxa}
\newsavebox{\teaserboxb}
\newsavebox{\teaserboxc}

\usepackage[breaklinks=true]{hyperref}
\usepackage{xspace}
\usepackage[font=small]{subcaption}
\usepackage{graphicx} 
\usepackage[dvipsnames]{xcolor}

\usepackage{multirow}
\usepackage{multicol}
\usepackage{soul}

\definecolor{blue(pigment)}{rgb}{0.2, 0.2, 0.9}
\hypersetup{colorlinks=true, citecolor=blue(pigment), linkcolor=blue(pigment), urlcolor=blue(pigment)}

\title{DLoop: Looped Speculative Decoding}

\author{%
  Geonmo Gu$^{1,\,3}$, Byeongho Heo$^{1}$, HeeJae Jun$^{2}$, Yoohoon Kang$^{2}$, Sangmin Lee$^{3}$, \\
  \bf Sangdoo Yun$^{1}$, Dongyoon Han$^{1}$ \\
  \normalfont $^{1}$NAVER AI Lab \qquad $^{2}$NAVER AI Search Platform \qquad $^{3}$Korea University
}

\iclrfinalcopy 
\begin{document}

\maketitle
\lhead{Preprint.}

\begin{abstract}
Speculative decoding accelerates autoregressive generation in large language models. In each drafting stage, a lightweight draft model proposes tokens that the target model subsequently verifies. With increasingly capable draft models, we find that the target model frequently accepts all tokens produced in a drafting stage. A verification nevertheless follows each drafting stage, resulting in unnecessary target-model forward passes even when drafting could have continued. Adaptive draft length methods decide during decoding how many draft tokens precede a verification, but they raise the speedup only for autoregressive draft models. For a parallel draft model, drafting further requires target-model hidden states for draft tokens that have not been verified. We propose \textbf{DLoop}, a looped form of speculative decoding that adaptively performs multiple drafting stages before verification. DLoop continues drafting while the draft model remains confident and verifies all accumulated draft tokens together. Loop-aware training keeps the draft model reliable in the additional drafting stages by exposing it to its own hidden states for unverified draft tokens. By spending additional draft-model forward passes, DLoop reduces the number of target-model forward passes required for verification. Across diverse speculative decoding methods including EAGLE-3, DFlash, Domino, DSpark, and multi-token prediction modules, DLoop improves the wall-clock speedup by 5 to 41 percent while preserving lossless decoding.
Code will be available at \href{https://github.com/naver-ai/DLoop}{\texttt{https://github.com/naver-ai/DLoop}}.

\end{abstract}

\section{Introduction}
Speculative decoding~\citep{leviathan2023fast, chen2023accelerating} accelerates autoregressive generation in large language models~\citep{brown2020language, achiam2023gpt, team2023gemini, grattafiori2024llama31, liu2024deepseek, yang2025qwen3}, where each token costs a full forward pass of the model. In each drafting stage, a lightweight draft model proposes several draft tokens that the target model is likely to generate next. It produces them either token by token~\citep{li2024eagle, li2024eagle2, li2025eagle3} or in parallel~\citep{chen2026dflash, cheng2026dspark, huang2026domino}. The target model then verifies these draft tokens in a single forward pass while preserving its output distribution.

\begin{wrapfigure}{r}{0.40\textwidth}
    \centering
    \includegraphics[width=\linewidth]{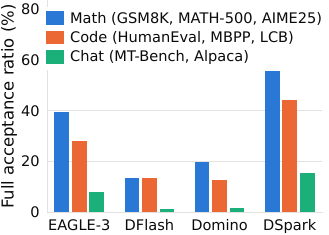}
    \caption{\textbf{Full acceptance is common.}
    Fraction of drafting stages in which the target model accepts every draft token, under standard speculative decoding with Qwen3-4B~\citep{yang2025qwen3} and greedy decoding ($T=0$).}
    \label{fig:full_acceptance}
    \vspace{-14.5pt}
\end{wrapfigure}

Speculative decoding performs a verification after every drafting stage, even when the draft model could have continued.
If the draft model keeps producing tokens that the target model accepts, additional lightweight drafting stages can replace costly verifications.
As shown in Fig.~\ref{fig:full_acceptance}, across four draft models and three task categories, a substantial fraction of drafting stages are fully accepted, that is, every draft token passes verification.
Verification also dominates decoding time, since it requires a full forward pass of the target model. The latency of that pass, however, grows only modestly with the number of tokens it verifies~\citep{pope2023efficiently}.

Adaptive draft length methods were proposed to address this problem by deciding during decoding how many draft tokens precede a verification~\citep{kim2023bild, zhang2025svip, agrawal2024adaedl, xu2023llmcad, huang2025specdec, mamou2024disco, zhang2026pacer, zhang2024adaeagle, brown2024ddd}.
These methods raise the speedup only for autoregressive draft models as shown in Section~\ref{sec:adaptive_comparison}. These methods build on an autoregressive draft model, either a separate small language model or a drafter of the EAGLE family~\citep{li2024eagle, li2024eagle2, li2025eagle3}. Such a draft model produces one token at a time and can therefore stop or continue drafting at any token. A parallel draft model~\citep{chen2026dflash, huang2026domino, cheng2026dspark} instead produces all draft tokens of a drafting stage in a single forward pass. Stopping early leaves the cost of this forward pass unchanged. Drafting further requires target-model hidden states for draft tokens that have not been verified. Without them, the draft model conditions on its own hidden states, a setting it was not trained for.

\begin{figure}[t]
\centering
    \includegraphics[width=\linewidth]{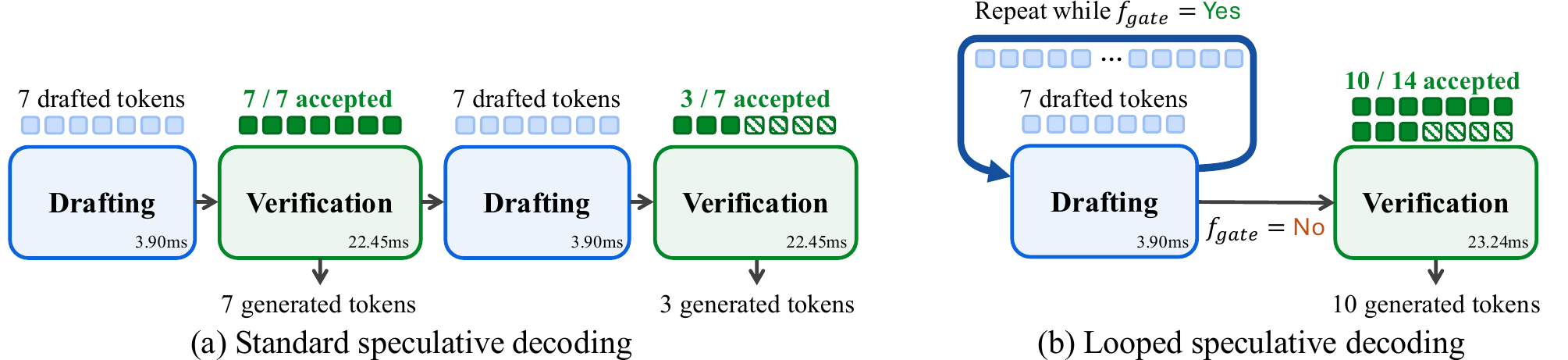}
    \caption{\textbf{Comparison of speculative and looped speculative decoding.}
    We assume that the target model accepts all 7 tokens of the first drafting stage and 3 of 7 from the second, 10 tokens in total.
    The bonus token that each verification produces is omitted for clarity.
    (a) Speculative decoding verifies after each drafting stage and needs two verifications.
    (b) Looped speculative decoding skips the first verification because the gate function $f_{\text{gate}}$ finds the first drafting stage confident. It then verifies all 14 draft tokens at once and yields the 10 tokens in a single verification.
    Latency is measured with a 1{,}024-token input prefix on a single H100 GPU, using Qwen3-4B~\citep{yang2025qwen3} as the target model and DSpark~\citep{cheng2026dspark} as the draft model.}
    \label{fig:overview}
    \vspace{-0.3cm}
\end{figure}

In this work, we propose \textbf{DLoop}, a looped form of speculative decoding in which a single verification covers multiple drafting stages. DLoop consists of two components, a \textit{confidence-gated drafting loop} at inference and \textit{loop-aware training}. The drafting loop, illustrated in Fig.~\ref{fig:overview}, decides how many drafting stages a single verification covers. It repeats the drafting stage while the draft model remains confident in the draft tokens it has just produced. After each drafting stage, the loop scores this confidence by summing the log probabilities of these draft tokens. If the score remains above a threshold, the loop performs an additional drafting stage. Otherwise, the target model verifies all accumulated draft tokens in one forward pass. Since the loop repeats whole drafting stages, it applies to parallel and autoregressive draft models alike. As in standard speculative decoding, the acceptance rule is unchanged and decoding remains lossless.

Loop-aware training aims to keep the draft model reliable in the additional drafting stages. An additional drafting stage conditions on the draft tokens of earlier drafting stages. Target-model hidden states exist only for the tokens before the first drafting stage. Loop-aware training exposes the draft model to its own hidden states for the draft tokens that follow. It is inspired by the training-time test of EAGLE-3~\citep{li2025eagle3}. That test unrolls the draft model over its own predicted features within one drafting stage. Loop-aware training extends this unrolling across the drafting stages of the loop. Unlike the training-time test, it thus applies to parallel draft models as well.

We evaluate DLoop on EAGLE-3~\citep{li2025eagle3}, DFlash~\citep{chen2026dflash}, Domino~\citep{huang2026domino}, DSpark~\citep{cheng2026dspark}, and multi-token prediction modules (MTP)~\citep{qwen35blog, team2026gemma4}. In every one of these settings, the loop and loop-aware training together raise both the acceptance length and the wall-clock speedup.
DLoop obtains this speedup by spending additional draft-model forward passes to reduce the number of target-model forward passes. It does not require additional parameters and applies on top of these publicly released draft models. DLoop does not change how a draft model produces tokens, only how many drafting stages precede a verification. The number of drafting stages before a verification need not be fixed by the schedule and can be decided during decoding.

\section{Related Work}
\label{sec:related_work}
\noindent\textbf{Speculative decoding.}
The speedup of speculative decoding~\citep{leviathan2023fast, chen2023accelerating} depends on the acceptance length and the cost of a drafting stage.
Early formulations train a separate draft model from the same family as the target model, whereas later work builds the draft model on the target model itself.
Decoding heads attached to the target model~\citep{stern2018blockwise, cai2024medusa} propose several future tokens in parallel from its already computed hidden states.
The EAGLE family~\citep{li2024eagle, li2024eagle2, li2025eagle3} replaces these parallel heads with a lightweight autoregressive drafter that conditions on the same hidden states.
The multi-token prediction modules released with recent models~\citep{qwen35blog, team2026gemma4} take the same autoregressive form, feeding each hidden state back to produce the next draft token.
EAGLE-3~\citep{li2025eagle3} trains its drafter with a training-time test, which unrolls it over its own predicted features within one drafting stage.
Loop-aware training extends that unrolling across the drafting stages that a single verification covers.
VAT~\citep{gu2026vat} instead reshapes the supervision itself, weighting each position by whether it survives a simulated verification.
The parallel draft models DFlash~\citep{chen2026dflash}, Domino~\citep{huang2026domino}, and DSpark~\citep{cheng2026dspark} remove the sequential cost of autoregressive drafting by producing a whole block in one forward pass.
Tree verification broadens what one drafting stage proposes to a tree of candidates that the target model verifies together~\citep{miao2024specinfer, li2024eagle2, ringel2026ddtree, li2026dartree}.
All of them change how a draft model produces a block, whereas DLoop changes only how many blocks precede a verification.
The same loop therefore runs on EAGLE-3, DFlash, Domino, DSpark, and the multi-token prediction modules.
It also composes with tree verification.

\noindent\textbf{Adaptive draft length.}
\label{sec:related_work_adaptive}
Most speculative decoding methods fix the length of a drafting stage in advance, whereas prior works have explored determining its length dynamically during decoding.
BiLD~\citep{kim2023bild} and Draft\&Verify~\citep{zhang2024draftverify} stop on the top-1 probability of the draft distribution.
SVIP~\citep{zhang2025svip} and AdaEDL~\citep{agrawal2024adaedl} stop on its entropy.
Dynamic depth decoding (DDD)~\citep{brown2024ddd} grows the draft tree~\citep{li2024eagle2} deeper while the summed probability of the sequences in the beam remains above a threshold.
LLMCad~\citep{xu2023llmcad} accumulates the confidences of the draft tokens within one drafting stage and calls the target model once the accumulated confidence falls below a threshold.
Our gate function also accumulates confidence over a block, but it decides whether a further drafting stage should follow.
Other methods train a module for the decision.
SpecDec++~\citep{huang2025specdec} predicts the acceptance probability of each draft token, DISCO~\citep{mamou2024disco} classifies whether to continue, and DSpark~\citep{cheng2026dspark} sets a load-aware verification length from calibrated acceptance probabilities.
AdaEAGLE~\citep{zhang2024adaeagle} predicts the length once, before the drafting stage begins.
All of them choose where the draft for one verification ends (for DSpark, how much of it is verified), whereas DLoop keeps a full drafting stage as the minimum and decides whether another drafting stage follows.
PACER~\citep{zhang2026pacer} drafts autoregressively in blocks and appends another block while a separately trained pre-verification layer's mean predicted acceptance for the latest block stays above a threshold.
Unlike PACER, DLoop gates on the confidence of the drafter itself and applies loop-aware training to the drafter.
SpecBlock~\citep{shi2026specblock} trains a new draft model that drafts up to a fixed number of blocks before a verification, and a co-trained rank head decides which positions start a later block.
DLoop instead applies to existing draft models without an additional module.

\section{Looped Speculative Decoding}
\label{sec:method}

\subsection{Background}
\label{sec:preliminaries}
\noindent\textbf{Preliminary.} Let $\mathcal{M}_t$ denote the target model, $\mathcal{M}_d$ the draft model, and $k$ the block size, the number of draft tokens proposed in one drafting stage. Let $S$ denote the loop depth, the number of drafting stages that a single verification covers. As shown in Fig.~\ref{fig:overview}(a), standard speculative decoding runs one drafting stage ($S=1$) and one verification. Drafting starts from a prefix $x_{1:t}$, the input prompt followed by the tokens verified so far, together with two quantities that $\mathcal{M}_t$ has produced for it. The first is the \emph{anchor token} $x_t$, the last token of the prefix, from which the next block is predicted. It is the token that $\mathcal{M}_t$ itself produced in the preceding forward pass, commonly referred to as the \emph{bonus token}. We call it the anchor token when a drafting stage starts from it and the bonus token when a verification produces it. The second is the target-side context, the hidden states $h_{1:t-1}$ gathered from a set of layers of $\mathcal{M}_t$, which $\mathcal{M}_d$ projects to its own hidden size. The context ends at position $t-1$ because $\mathcal{M}_t$ has generated the anchor token by next-token prediction but has not yet run a forward pass on it.

In the drafting stage, $\mathcal{M}_d$ conditions on the two quantities and proposes a block of $k$ draft tokens,
\begin{equation}
\hat{x}_{t+1:t+k} = \mathcal{M}_d(h_{1:t-1},\, x_t),
\end{equation}
each with a log probability under $\mathcal{M}_d$. Existing draft methods differ in how the context is injected and in how the $k$ tokens are produced. An autoregressive draft model~\citep{li2025eagle3, qwen35blog, team2026gemma4} produces them one at a time and drafts each token after the first from its own hidden states. A parallel draft model~\citep{chen2026dflash, huang2026domino, cheng2026dspark} instead produces all $k$ tokens in a single forward pass. 

In the verification, $\mathcal{M}_t$ processes the block in a single forward pass and applies the acceptance rule of speculative sampling~\citep{leviathan2023fast, chen2023accelerating}. If the first $a \in \{0,\dots,k\}$ draft tokens are accepted, the verification adds these $a$ tokens to the prefix together with a bonus token produced by $\mathcal{M}_t$. The hidden states of these positions come from the same pass. The token produced by $\mathcal{M}_t$ becomes the anchor token of the next drafting stage.

\noindent\textbf{Limitation of fixed verification procedure.} We argue that the de facto standard draft-verify procedure is inherently inefficient due to its fixed alternating structure. Specifically, 
even when all $k$ draft tokens are accepted, the next drafting stage must wait for another verification. Thus, one verification yields at most $k+1$ tokens regardless of how accurately $\mathcal{M}_d$ drafts. Fig.~\ref{fig:full_acceptance} shows that a substantial fraction of blocks reach this bound. This bound is costly because verification requires a full forward pass of the target model and dominates decoding time.
Adaptive draft length methods~\citep{kim2023bild, zhang2025svip, huang2025specdec, cheng2026dspark} decide during decoding how many draft tokens precede a verification. Section~\ref{sec:adaptive_comparison} shows that they raise the speedup only for autoregressive draft models. For a parallel draft model, stopping early leaves the cost of a drafting stage unchanged. Drafting further requires target-model hidden states for draft tokens that have not been verified. Our goal is to allow a single verification to cover multiple drafting stages.

\subsection{Confidence-Gated Drafting Loop}
\label{sec:loop}

DLoop performs $S \geq 1$ drafting stages and a single verification, as shown in Fig.~\ref{fig:overview}(b). The $s$-th drafting stage produces a block $\hat{x}_{t+(s-1)k+1:t+sk}$. After each drafting stage, a gate function decides whether to perform another. When the gate function stops the loop, the target model verifies the accumulated draft tokens $\hat{x}_{t+1:t+Sk}$ in one forward pass. The gate function decides the number of drafting stages $S$ during decoding. Since the loop repeats whole drafting stages, it applies to parallel and autoregressive draft models alike. Under full acceptance, one verification yields at most $Sk+1$ tokens, whereas $S$ separate verifications yield at most $S(k+1)$. DLoop therefore forgoes up to $S-1$ intermediate bonus-token opportunities in exchange for the $S-1$ target-model forward passes it does not perform.

\noindent\textbf{Gate function.}
Whether a block is fully accepted is known only after verification. The gate function therefore predicts from the draft distribution whether the latest block will be fully accepted. For each draft token $\hat{x}_i$ of the block just produced, we take the log probability $\ell_i$ that $\mathcal{M}_d$ assigns to it. Since the probability is normalized over the vocabulary, $\ell_i$ measures how much mass $\mathcal{M}_d$ places on $\hat{x}_i$ among all tokens at that position.
The score $c_s$ of the $s$-th block is the sum of these log probabilities, and the gate function compares it with a threshold $g$,
\begin{equation}
c_s = \sum_{i=t+(s-1)k+1}^{t+sk} \ell_i ,
\qquad
f_{\text{gate}}(c_s) =
\begin{cases}
1 & \text{if } c_s \geq g, \\
0 & \text{otherwise}.
\end{cases}
\end{equation}
The sum corresponds to the product of the per-token probabilities, a quantity prior work uses to rank draft candidates~\citep{li2024eagle2} and to set the depth of a drafting stage~\citep{brown2024ddd}. Since every $\ell_i$ is non-positive, a confident position cannot offset an uncertain one. The sum therefore falls whenever a single position is uncertain, in line with the acceptance rule that discards every token after the first rejection. DLoop performs another drafting stage while $f_{\text{gate}}(c_s) = 1$. Because the log probabilities $\ell_i$ are already computed in the drafting stage, the gate function adds only the summation and one comparison, without an additional module or forward pass. The score considers only the latest block, since reaching the $s$-th drafting stage already implies that every preceding block passed the gate. The threshold $g$ is the only hyperparameter of the loop and is fixed before decoding.

\noindent\textbf{Verification.}
The verification applies the acceptance rule of Section~\ref{sec:preliminaries} to the accumulated draft tokens as if they were a single block of $Sk$ tokens. Since the rule preserves the output distribution of $\mathcal{M}_t$ for a draft sequence of any length, decoding remains lossless~\citep{leviathan2023fast, chen2023accelerating}. If a token in an earlier block is rejected, the verification also discards every later block. The drafting stages that produced them then add cost without adding accepted tokens. The accepted positions receive their hidden states $h$ from the same pass. The next loop starts from the new anchor token as in standard speculative decoding.

\noindent\textbf{Input to the additional drafting stages.}
The first drafting stage takes the target-side context $h_{1:t-1}$ and the anchor token $x_t$, as in Section~\ref{sec:preliminaries}. The additional drafting stages lack this input, since the draft tokens $\hat{x}_{t+1:t+(s-1)k}$ are not yet verified and $\mathcal{M}_t$ has not computed hidden states for them. We instead use the hidden states of $\mathcal{M}_d$. While producing a block, $\mathcal{M}_d$ computes its own final-layer hidden states $\tilde{h}$ at the positions it has already processed. The last draft token of the previous block takes the role of the anchor token for the $s$-th drafting stage. For $s \geq 2$,
\begin{equation}
\hat{x}_{t+(s-1)k+1:t+sk}
= \mathcal{M}_d\big(
h_{1:t-1} \mathbin{\Vert} \tilde{h}_{t:t+(s-1)k-1},
\, \hat{x}_{t+(s-1)k}
\big),
\label{eq:loop_input}
\end{equation}
where $\Vert$ denotes concatenation along the position dimension.
The context is the target-side hidden states of the prefix followed by the draft-side hidden states from the first anchor token onward. It ends one position before the anchor token of the $s$-th drafting stage, as in standard speculative decoding. The hidden states $\tilde{h}$ occupy the positions that $h$ would fill after verification. Each drafting stage therefore processes only its $k$ new positions. The anchor token of the first drafting stage is produced by $\mathcal{M}_t$ and is always kept. The anchor token of an additional drafting stage is a draft token that the gate function has judged confident, but $\mathcal{M}_t$ has not yet verified. A parallel draft model~\citep{chen2026dflash, huang2026domino, cheng2026dspark}, however, never receives $\tilde{h}$ as input during training. An autoregressive draft model~\citep{li2025eagle3, qwen35blog, team2026gemma4} receives $\tilde{h}$ only within a block. Section~\ref{sec:training} describes how we train the draft model to take $\tilde{h}$ as input.

\subsection{Loop-Aware Training}
\label{sec:training}
The draft models~\citep{li2025eagle3, chen2026dflash, cheng2026dspark, huang2026domino, qwen35blog, team2026gemma4} we build on are trained for a single drafting stage before each verification. During training, the first drafting stage receives the hidden states $h_{1:t-1}$ that $\mathcal{M}_t$ computed on a prefix and is supervised to predict the $k$ tokens that follow. EAGLE-3~\citep{li2025eagle3} unrolls this supervision over the draft model's own hidden states for the depth of one drafting stage. Loop-aware training starts from that design. Each additional drafting stage in Section~\ref{sec:loop} starts from a chain of $\tilde{h}$ that runs past that depth. Loop-aware training extends the unroll to the chain that one verification covers.

\paragraph{Training scheme.}
For each training sequence and position $t$, we sample the number of additional drafting stages $S - 1$ uniformly from $\{1, \dots, N\}$ and run $S$ drafting stages. The gate function decides $S$ during decoding. Training at one length would therefore expose the draft model to a single point in the range it meets. The first drafting stage takes $h_{1:t-1}$ from $\mathcal{M}_t$. Each additional drafting stage takes the chain of final-layer hidden states $\tilde{h}$ that $\mathcal{M}_d$ computed in the preceding drafting stages, as in the loop. The tokens at these positions come from the ground truth. Training therefore reproduces the chain of hidden states that the loop builds and not the draft tokens that accompany it. To simulate the drafting process in Eq.~\ref{eq:loop_input}, we define $\mathcal{L}^{(s)}$ as the cross-entropy of the $s$-th drafting stage against the ground-truth tokens $x_{t+(s-1)k+1:t+sk}$. The training loss combines the first and the last drafting stage,
\begin{equation}
\mathcal{L} = \frac{1}{2}\left(\mathcal{L}^{(1)} + \mathcal{L}^{(S)}\right).
\label{eq:total_loss}
\end{equation}

The first term is the original objective and preserves the first drafting stage. The second term trains the draft model to continue from the previous $\tilde{h}$. The additional drafting stages differ in difficulty. The intermediate drafting stages predict nearer blocks and form the easier part of the loop. The last drafting stage predicts the farthest block from the longest chain of $\tilde{h}$ and forms the hardest part. We place the second loss term only on the last drafting stage. Every training iteration therefore focuses on this hardest part. The intermediate drafting stages do not contribute to the loss and only supply $\tilde{h}$ to the next drafting stage. Their logits are never computed. Loop-aware training does not require an additional forward pass of the target model, since $h$ is used only for the prefix. The draft model instead runs $S$ drafting stages per training iteration. Loop-aware training leaves the architecture and the parameter count unchanged and applies identically to every draft model on top of its publicly released weights. Tab.~\ref{tab:kmax_midce} ablates $N$ and reports the training time and memory. We use $N = 3$ in our experiments.

\section{Experiment}

\subsection{Experimental Setup}
\label{sec:setup}
We evaluate DLoop on four draft models, EAGLE-3~\citep{li2025eagle3}, DFlash~\citep{chen2026dflash}, Domino~\citep{huang2026domino}, and DSpark~\citep{cheng2026dspark}, with Qwen3-4B and Qwen3-8B~\citep{yang2025qwen3} as target models. We also evaluate DLoop with the MTP modules released with Qwen3.5-9B~\citep{qwen35blog} and Gemma-4-12B~\citep{team2026gemma4}. The block size is 16 for DFlash and Domino, 7 for DSpark, 3 for MTP with Qwen3.5-9B, and 6 for MTP with Gemma-4-12B. Loop-aware training starts from the released official weights of each draft model and runs for one epoch. Following prior works~\citep{li2025eagle3, chen2026dflash, huang2026domino, cheng2026dspark}, we construct the training set from the 1.3 million user prompts of PerfectBlend~\citep{xu2024perfectblend} paired with responses regenerated by each target model with its recommended sampling parameters. All training hyperparameters follow those released with each method. Multi-token prediction (MTP) methods~\citep{qwen35blog, team2026gemma4} do not provide a recipe for training only the MTP modules. For these methods, we therefore use the training recipe of EAGLE-3. EAGLE-3 has no officially released weights for these target models. We therefore train EAGLE-3 draft models for three epochs on the same data and use them as the baseline draft model.

We follow the evaluation protocol of DFlash\footnote{\url{https://github.com/z-lab/dflash}}, detailed in Appendix~\ref{app:protocol}, and evaluate on eight benchmarks in three task categories, math (GSM8K~\citep{cobbe2021gsm8k}, MATH-500~\mbox{\citep{lightman2023math500}}, AIME25), code (HumanEval~\citep{chen2021humaneval}, MBPP~\citep{austin2021mbpp}, LiveCodeBench (LCB)~\citep{jain2025livecodebench}), and chat (MT-Bench~\citep{zheng2023mtbench}, Alpaca~\citep{taori2023alpaca}). We report the acceptance length $\tau$, the average number of tokens accepted per verification, and the wall-clock speedup over autoregressive decoding. All results use greedy decoding ($T=0$) unless stated otherwise. For tree verification, we apply DDTree~\citep{ringel2026ddtree} to DFlash and DARTree~\citep{li2026dartree} to Domino and DSpark. We follow the maximum number of node budget in the tree that the target model verifies from their papers: 256 for DFlash, 64 for Domino and DSpark. EAGLE-3 uses its own tree verification with the released setting. We sweep the gate threshold $g$ over $\{-0.25, -0.5, -0.75, -1.0\}$, as detailed in the appendix Tab.~\ref{tab:gate_sweep} and select the value with the highest average speedup over the eight benchmarks. We use $-0.5$ for all settings, except EAGLE-3 with Qwen3-8B, where we use $-0.75$. All training and evaluation runs use NVIDIA H100 80GB GPUs with bf16 precision.

\begin{table}[t]
\centering
\caption{\textbf{Decoding speedup over the autoregressive baseline and average acceptance length ($\tau$).}
Results are averaged over the benchmarks in each task category and over all eight benchmarks (Avg.). For each draft model, the first row is the publicly released draft model. ``$+$~DLoop'' applies one epoch of loop-aware training and enables the confidence-gated drafting loop. Bold marks the better result for each draft model. Per-benchmark results are given in Table~\ref{tab:full} in the appendix.}
\label{tab:main}
\setlength{\tabcolsep}{2.5pt}
\renewcommand{\arraystretch}{1.0}
\resizebox{\textwidth}{!}{\footnotesize
\begin{tabular}{@{}l*{4}{r}c*{4}{r}@{}}
\toprule
 & \multicolumn{4}{c}{\textbf{Qwen3-4B}} & & \multicolumn{4}{c}{\textbf{Qwen3-8B}} \\
\cmidrule(lr){2-5} \cmidrule(l){7-10}
\multirow{2}{*}{Method} & \multicolumn{1}{c}{\textsc{Math}} & \multicolumn{1}{c}{\textsc{Code}} & \multicolumn{1}{c}{\textsc{Chat}} & \multicolumn{1}{c}{\textit{Avg.}} & & \multicolumn{1}{c}{\textsc{Math}} & \multicolumn{1}{c}{\textsc{Code}} & \multicolumn{1}{c}{\textsc{Chat}} & \multicolumn{1}{c}{\textit{Avg.}} \\
\cmidrule(lr){2-2} \cmidrule(lr){3-3} \cmidrule(lr){4-4} \cmidrule(lr){5-5} \cmidrule(lr){7-7} \cmidrule(lr){8-8} \cmidrule(lr){9-9} \cmidrule(lr){10-10}
 & $\tau$\,/\,spd & $\tau$\,/\,spd & $\tau$\,/\,spd & $\tau$\,/\,spd & & $\tau$\,/\,spd & $\tau$\,/\,spd & $\tau$\,/\,spd & $\tau$\,/\,spd \\
\midrule
EAGLE-3~\citeyear{li2025eagle3}                          & 7.83\,/\,3.85$\times$ & 7.09\,/\,3.49$\times$ & 5.30\,/\,2.38$\times$ & 6.92\,/\,3.35$\times$ & & 7.59\,/\,3.78$\times$ & 6.73\,/\,3.30$\times$ & 4.98\,/\,2.26$\times$ & 6.61\,/\,3.22$\times$ \\
\;\;$+$ \textbf{DLoop}           & \textbf{12.04}\,/\,\textbf{4.58}$\times$ & \textbf{9.43}\,/\,\textbf{3.83}$\times$ & \textbf{6.29}\,/\,\textbf{2.47}$\times$ & \textbf{9.63}\,/\,\textbf{3.77}$\times$ & & \textbf{10.45}\,/\,\textbf{4.31}$\times$ & \textbf{7.83}\,/\,\textbf{3.46}$\times$ & \textbf{5.46}\,/\,\textbf{2.38}$\times$ & \textbf{8.22}\,/\,\textbf{3.51}$\times$ \\
\midrule
DFlash~\citeyear{chen2026dflash}                           & 7.19\,/\,5.02$\times$ & 6.59\,/\,4.65$\times$ & 3.74\,/\,2.27$\times$ & 6.10\,/\,4.19$\times$ & & 7.22\,/\,5.13$\times$ & 6.56\,/\,4.45$\times$ & 3.66\,/\,2.22$\times$ & 6.08\,/\,4.15$\times$ \\
\;\;$+$ \textbf{DLoop}           & \textbf{9.48}\,/\,\textbf{6.14}$\times$ & \textbf{6.99}\,/\,\textbf{4.72}$\times$ & \textbf{4.12}\,/\,\textbf{2.38}$\times$ & \textbf{7.21}\,/\,\textbf{4.67}$\times$ & & \textbf{9.78}\,/\,\textbf{6.28}$\times$ & \textbf{7.21}\,/\,\textbf{4.70}$\times$ & \textbf{4.08}\,/\,\textbf{2.37}$\times$ & \textbf{7.40}\,/\,\textbf{4.71}$\times$ \\
\midrule
Domino~\citeyear{huang2026domino}                           & 8.85\,/\,5.63$\times$ & 6.99\,/\,4.39$\times$ & 4.56\,/\,2.48$\times$ & 7.08\,/\,4.38$\times$ & & 9.04\,/\,5.75$\times$ & 7.18\,/\,4.49$\times$ & 4.54\,/\,2.51$\times$ & 7.22\,/\,4.47$\times$ \\
\;\;$+$ \textbf{DLoop}           & \textbf{10.44}\,/\,\textbf{5.98}$\times$ & \textbf{7.41}\,/\,\textbf{4.54}$\times$ & \textbf{4.82}\,/\,\textbf{2.55}$\times$ & \textbf{7.90}\,/\,\textbf{4.58}$\times$ & & \textbf{11.14}\,/\,\textbf{6.31}$\times$ & \textbf{7.69}\,/\,\textbf{4.64}$\times$ & \textbf{4.91}\,/\,\textbf{2.61}$\times$ & \textbf{8.29}\,/\,\textbf{4.76}$\times$ \\
\midrule
DSpark~\citeyear{cheng2026dspark}                           & 6.26\,/\,4.36$\times$ & 5.61\,/\,3.96$\times$ & 4.09\,/\,2.62$\times$ & 5.47\,/\,3.77$\times$ & & 6.32\,/\,4.46$\times$ & 5.81\,/\,3.96$\times$ & 4.13\,/\,2.61$\times$ & 5.58\,/\,3.81$\times$ \\
\;\;$+$ \textbf{DLoop}           & \textbf{11.02}\,/\,\textbf{6.02}$\times$ & \textbf{8.44}\,/\,\textbf{4.86}$\times$ & \textbf{5.17}\,/\,\textbf{2.71}$\times$ & \textbf{8.59}\,/\,\textbf{4.76}$\times$ & & \textbf{11.80}\,/\,\textbf{6.20}$\times$ & \textbf{8.84}\,/\,\textbf{4.83}$\times$ & \textbf{5.39}\,/\,\textbf{2.74}$\times$ & \textbf{9.08}\,/\,\textbf{4.82}$\times$ \\
\midrule
 & \multicolumn{4}{c}{\textbf{Qwen3.5-9B}} & & \multicolumn{4}{c}{\textbf{Gemma-4-12B}} \\
\cmidrule(lr){2-5} \cmidrule(l){7-10}
MTP~\citeyear{team2026qwen3.5, team2026gemma4}                              & 3.54\,/\,1.74$\times$ & 3.39\,/\,1.58$\times$ & 2.88\,/\,1.21$\times$ & 3.32\,/\,1.55$\times$ & & 5.38\,/\,3.56$\times$ & 5.12\,/\,3.37$\times$ & 3.71\,/\,2.28$\times$ & 4.86\,/\,3.17$\times$ \\
\;\;$+$ \textbf{DLoop}           & \textbf{8.26}\,/\,\textbf{2.66}$\times$ & \textbf{6.92}\,/\,\textbf{2.26}$\times$ & \textbf{4.02}\,/\,\textbf{1.33}$\times$ & \textbf{6.70}\,/\,\textbf{2.18}$\times$ & & \textbf{9.06}\,/\,\textbf{4.33}$\times$ & \textbf{8.01}\,/\,\textbf{3.94}$\times$ & \textbf{4.40}\,/\,\textbf{2.30}$\times$ & \textbf{7.50}\,/\,\textbf{3.68}$\times$ \\
\bottomrule
\end{tabular}}
\end{table}
\subsection{Experimental Results}
\noindent\textbf{DLoop on recent draft models}
\label{sec:main_results}
Table~\ref{tab:main} reports the results for the four draft models with Qwen3-4B and Qwen3-8B, and for MTP with Qwen3.5-9B and Gemma-4-12B. Applying DLoop to the publicly released draft models keeps their architecture and block size. DLoop lets the draft model continue drafting after a block instead of verifying each block on its own. This improves both the average acceptance length and the wall-clock speedup for every draft model, target model, and task category. Across these settings, the average acceptance length increases by 12 to 102 percent and the wall-clock speedup by 5 to 41 percent. Among the four draft models with Qwen3-4B and Qwen3-8B, the largest gain is observed for DSpark. DSpark also has the highest full acceptance ratio in Fig.~\ref{fig:full_acceptance} and thus the most blocks after which an additional drafting stage can be accepted. Results at $T=1$ are reported in Tab.~\ref{tab:full_t1}.

\begin{table}[t]
\centering
\caption{\textbf{Ablation of the confidence-gated drafting loop and loop-aware training.}
Each cell reports the average acceptance length ($\tau$) and speedup (spd) over all eight benchmarks as $\tau$\,/\,spd$\times$. ``Loop'' enables the confidence-gated drafting loop, and ``Train'' applies one epoch of loop-aware training. The first row is the publicly released draft model, and the last row is DLoop in Table~\ref{tab:main}. Q4B, Q8B, and Q9B denote Qwen3-4B, Qwen3-8B, and Qwen3.5-9B, and G12B denotes Gemma-4-12B. Bold marks the best result in each column. Per-benchmark results are given in Table~\ref{tab:full} in the appendix.}
\label{tab:dloop_ablation}
\setlength{\tabcolsep}{2.5pt}
\renewcommand{\arraystretch}{1.0}
\resizebox{\textwidth}{!}{\footnotesize
\begin{tabular}{@{}cc*{10}{c}@{}}
\toprule
\multirow{2}{*}{\textbf{Loop}} & \multirow{2}{*}{\textbf{Train}} & \multicolumn{2}{c}{\textbf{EAGLE-3}} & \multicolumn{2}{c}{\textbf{DFlash}} & \multicolumn{2}{c}{\textbf{Domino}} & \multicolumn{2}{c}{\textbf{DSpark}} & \multicolumn{2}{c}{\textbf{MTP}} \\
\cmidrule(lr){3-4} \cmidrule(lr){5-6} \cmidrule(lr){7-8} \cmidrule(lr){9-10} \cmidrule(l){11-12}
 & & Q4B & Q8B & Q4B & Q8B & Q4B & Q8B & Q4B & Q8B & Q9B & G12B \\
\midrule
           &            & 6.92\,/\,3.35$\times$ & 6.61\,/\,3.22$\times$ & 6.10\,/\,4.19$\times$ & 6.08\,/\,4.15$\times$ & 7.08\,/\,4.38$\times$ & 7.22\,/\,4.47$\times$ & 5.47\,/\,3.77$\times$ & 5.58\,/\,3.81$\times$ & 3.32\,/\,1.55$\times$ & 4.86\,/\,3.17$\times$ \\
\checkmark &            & 8.04\,/\,3.52$\times$ & 7.34\,/\,3.38$\times$ & 6.11\,/\,4.16$\times$ & 6.20\,/\,4.17$\times$ & 7.15\,/\,4.32$\times$ & 7.40\,/\,4.51$\times$ & 5.77\,/\,3.79$\times$ & 5.92\,/\,3.87$\times$ & 4.96\,/\,1.80$\times$ & 6.06\,/\,2.90$\times$ \\
           & \checkmark & 6.99\,/\,3.37$\times$ & 6.60\,/\,3.15$\times$ & 6.38\,/\,4.23$\times$ & 6.48\,/\,4.33$\times$ & 7.06\,/\,4.38$\times$ & 7.23\,/\,4.52$\times$ & 5.41\,/\,3.73$\times$ & 5.50\,/\,3.77$\times$ & 3.37\,/\,1.60$\times$ & 4.95\,/\,3.22$\times$ \\
\checkmark & \checkmark & \textbf{9.63}\,/\,\textbf{3.77}$\times$ & \textbf{8.22}\,/\,\textbf{3.51}$\times$ & \textbf{7.21}\,/\,\textbf{4.67}$\times$ & \textbf{7.40}\,/\,\textbf{4.71}$\times$ & \textbf{7.90}\,/\,\textbf{4.58}$\times$ & \textbf{8.29}\,/\,\textbf{4.76}$\times$ & \textbf{8.59}\,/\,\textbf{4.76}$\times$ & \textbf{9.08}\,/\,\textbf{4.82}$\times$ & \textbf{6.70}\,/\,\textbf{2.18}$\times$ & \textbf{7.50}\,/\,\textbf{3.68}$\times$ \\
\bottomrule
\end{tabular}}
\vspace{-0.3cm}
\end{table}

\noindent\textbf{Ablation of DLoop components}
\label{sec:dloop_ablation}
To separate the two components of DLoop, we enable the confidence-gated drafting loop and loop-aware training one at a time in Tab.~\ref{tab:dloop_ablation}. Enabling the loop without loop-aware training increases the average acceptance length for every draft model and target model. For the autoregressive draft models, which already draft each token after the first from their own hidden states, both the acceptance length and the speedup improve for EAGLE-3 with both target models and for MTP with Qwen3.5-9B. In the other settings, the speedup changes little or decreases, since the time added by the additional drafting stages largely offsets the gain in acceptance length. Loop-aware training alone changes both metrics by less than 7 percent, leaving them close to those of the publicly released draft models. Combined with the loop, however, it improves both metrics in every setting, and the gain over the publicly released draft model exceeds the sum of the gains from the loop alone and the training alone. Loop-aware training thus prepares the draft model for the loop rather than improving each drafting stage on its own. Detailed design choices for the gate function and loop-aware training are provided in Appendix~\ref{app:design_choices}.

\begin{table}[t]
\centering
\caption{\textbf{Comparison with adaptive draft length methods.}
All methods are applied to EAGLE-3, an autoregressive draft model, and DSpark, a parallel draft model, with Qwen3-4B. Each cell reports the average acceptance length ($\tau$) and speedup (spd) as $\tau$\,/\,spd$\times$, averaged over the benchmarks in each task category and over all eight benchmarks (Avg.). Cells without results are methods that are not applicable to the draft model, as discussed in Section~\ref{sec:adaptive_comparison}. The first row is the publicly released draft model, and ``$+$~DLoop'' is the same as in Table~\ref{tab:main}. Bold marks the best result in each column, and underline marks results higher than the draft model.}
\label{tab:adaptive}
\setlength{\tabcolsep}{2.5pt}
\renewcommand{\arraystretch}{1.0}
\resizebox{\textwidth}{!}{\footnotesize
\begin{tabular}{@{}l*{4}{c}c*{4}{c}@{}}
\toprule
 & \multicolumn{4}{c}{\textbf{EAGLE-3}} & & \multicolumn{4}{c}{\textbf{DSpark}} \\
\cmidrule(lr){2-5} \cmidrule(l){7-10}
Method & \textsc{Math} & \textsc{Code} & \textsc{Chat} & \textit{Avg.} & & \textsc{Math} & \textsc{Code} & \textsc{Chat} & \textit{Avg.} \\
\midrule
Draft model                                        & 7.83\,/\,3.85$\times$ & 7.09\,/\,3.49$\times$ & 5.30\,/\,2.38$\times$ & 6.92\,/\,3.35$\times$ & & 6.26\,/\,4.36$\times$ & 5.61\,/\,3.96$\times$ & 4.09\,/\,2.62$\times$ & 5.47\,/\,3.77$\times$ \\
\;\;$+$ BiLD~\citep{kim2023bild}                   & \underline{8.70}\,/\,\underline{4.01}$\times$ & \underline{7.63}\,/\,3.49$\times$ & \underline{5.31}\,/\,2.31$\times$ & \underline{7.45}\,/\,\underline{3.39}$\times$ & & 6.26\,/\,4.31$\times$ & 5.61\,/\,3.93$\times$ & 4.09\,/\,2.60$\times$ & 5.47\,/\,3.74$\times$ \\
\;\;$+$ SVIP~\citep{zhang2025svip}                 & 7.73\,/\,\underline{3.86}$\times$ & 7.06\,/\,\underline{3.51}$\times$ & 5.10\,/\,\underline{2.40}$\times$ & 6.82\,/\,\underline{3.36}$\times$ & & 6.08\,/\,4.17$\times$ & 5.56\,/\,3.90$\times$ & 3.94\,/\,2.47$\times$ & 5.35\,/\,3.65$\times$ \\
\;\;$+$ AdaEDL~\citep{agrawal2024adaedl}           & 7.83\,/\,3.84$\times$ & 7.09\,/\,3.46$\times$ & 5.20\,/\,2.37$\times$ & 6.90\,/\,3.33$\times$ & & 6.25\,/\,4.30$\times$ & 5.61\,/\,3.93$\times$ & 4.08\,/\,2.57$\times$ & 5.47\,/\,3.73$\times$ \\
\;\;$+$ LLMCad~\citep{xu2023llmcad}                & \underline{9.18}\,/\,\underline{4.10}$\times$ & \underline{7.40}\,/\,\underline{3.53}$\times$ & 5.09\,/\,\textbf{\underline{2.51}}$\times$ & \underline{7.49}\,/\,\underline{3.49}$\times$ & & 6.11\,/\,4.18$\times$ & 5.45\,/\,3.79$\times$ & 3.69\,/\,2.29$\times$ & 5.26\,/\,3.56$\times$ \\
\;\;$+$ SpecDec++~\citep{huang2025specdec}         & \underline{8.85}\,/\,\underline{3.87}$\times$ & \underline{7.44}\,/\,3.40$\times$ & 5.04\,/\,\underline{2.43}$\times$ & \underline{7.37}\,/\,3.33$\times$ & & \underline{6.27}\,/\,4.32$\times$ & 5.60\,/\,3.91$\times$ & 4.05\,/\,2.56$\times$ & 5.46\,/\,3.73$\times$ \\
\;\;$+$ DISCO~\citep{mamou2024disco}               & 7.71\,/\,3.67$\times$ & 7.04\,/\,3.35$\times$ & 5.10\,/\,2.38$\times$ & 6.81\,/\,3.23$\times$ & & \underline{6.27}\,/\,4.31$\times$ & 5.61\,/\,3.88$\times$ & 4.09\,/\,2.57$\times$ & \underline{5.48}\,/\,3.71$\times$ \\
\;\;$+$ PACER~\citep{zhang2026pacer}               & \underline{9.08}\,/\,3.84$\times$ & \underline{7.66}\,/\,3.28$\times$ & \underline{5.32}\,/\,2.35$\times$ & \underline{7.61}\,/\,3.26$\times$ & & \underline{6.85}\,/\,4.22$\times$ & \underline{5.92}\,/\,3.75$\times$ & \underline{4.19}\,/\,2.53$\times$ & \underline{5.83}\,/\,3.62$\times$ \\
\;\;$+$ AdaEAGLE~\citep{zhang2024adaeagle}         & 7.71\,/\,\underline{3.86}$\times$ & 6.94\,/\,\underline{3.51}$\times$ & 4.83\,/\,\underline{2.46}$\times$ & 6.70\,/\,\underline{3.38}$\times$ & & 6.23\,/\,4.30$\times$ & 5.55\,/\,3.87$\times$ & 3.79\,/\,2.37$\times$ & 5.37\,/\,3.66$\times$ \\
\;\;$+$ DDD~\citep{brown2024ddd}                   & \underline{8.76}\,/\,\underline{4.20}$\times$ & \underline{7.71}\,/\,\underline{3.73}$\times$ & \underline{5.38}\,/\,2.41$\times$ & \underline{7.52}\,/\,\underline{3.58}$\times$ & & \underline{6.71}\,/\,\underline{4.39}$\times$ & \underline{5.89}\,/\,3.94$\times$ & \underline{4.19}\,/\,2.59$\times$ & \underline{5.77}\,/\,3.77$\times$ \\
\;\;$+$ DSpark's confidence scheduler              & \multicolumn{4}{c}{\emph{specific to DSpark}} & & 6.26\,/\,\underline{4.38}$\times$ & 5.61\,/\,3.87$\times$ & 4.09\,/\,\underline{2.63}$\times$ & 5.47\,/\,3.75$\times$ \\
\;\;$+$ \textbf{DLoop}                             & \textbf{12.04}\,/\,\textbf{4.58}$\times$ & \textbf{9.43}\,/\,\textbf{3.83}$\times$ & \textbf{6.29}\,/\,\underline{2.47}$\times$ & \textbf{9.63}\,/\,\textbf{3.77}$\times$ & & \textbf{11.02}\,/\,\textbf{6.02}$\times$ & \textbf{8.44}\,/\,\textbf{4.86}$\times$ & \textbf{5.17}\,/\,\textbf{2.71}$\times$ & \textbf{8.59}\,/\,\textbf{4.76}$\times$ \\
\bottomrule
\end{tabular}}
\end{table}

\noindent\textbf{Comparison with adaptive draft length methods.} 
\label{sec:adaptive_comparison}
The adaptive draft length methods in Section~\ref{sec:related_work_adaptive} also decide during decoding how many draft tokens precede a verification. BiLD, SVIP, AdaEDL, LLMCad, and DDD stop drafting once a statistic of the draft distribution, such as the top-1 probability or the entropy, crosses a threshold. SpecDec++ and DISCO instead train a head that decides after each draft token whether to stop, whereas AdaEAGLE trains a predictor that sets the draft length in advance. PACER trains a pre-verification layer and drafts another block while the acceptance this layer predicts remains above a threshold. The confidence scheduler of DSpark chooses how many draft tokens of the full block to verify, using a confidence head trained together with DSpark. BiLD, SVIP, AdaEDL, LLMCad, SpecDec++, DISCO, and PACER were proposed for a separate small autoregressive LLM, whose draft length can be varied freely because it is a standalone language model. In contrast, EAGLE-3 and DSpark are trained for a \textit{fixed draft length}, and drafting beyond this length quickly loses efficiency. We apply these methods to EAGLE-3 and DSpark as detailed in Appendix~\ref{app:adaptive_baselines}. Because the parallel draft model DSpark computes all $k$ draft tokens of a drafting stage in a single forward pass, cutting its draft shortens only the verification. On DSpark, PACER and DDD instead decide at the end of each drafting stage whether another drafting stage follows. We train the heads of SpecDec++, DISCO, AdaEAGLE, and PACER for one epoch on the same data as loop-aware training. Tab.~\ref{tab:adaptive} shows that on DSpark no method exceeds the speedup of the draft model, and on EAGLE-3 the seven methods above raise it by at most 0.14$\times$, as their stopping decisions add computation to the drafting loop. With loop-aware training for the additional drafting stages, DLoop gives the highest average acceptance length and speedup on both draft models.

\begin{figure}[t]
\centering
    \includegraphics[width=1.0\linewidth]{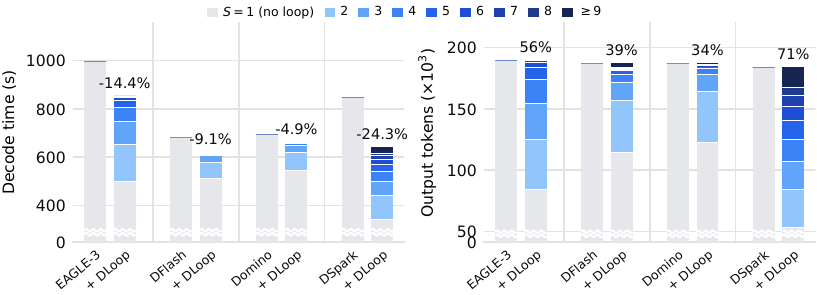}
    \caption{\textbf{Decode time and output tokens by the number of drafting stages.}
We break down the total decode time (left) and output tokens (right) by the number of drafting stages $S$ in each loop, summed over the math benchmarks with Qwen3-4B.
The percentages above the left bars show the reduction in total decode time from DLoop, while those above the right bars show the fraction of output tokens generated in loops with $S>1$.}
    \label{fig:loop_breakdown}
    \vspace{-0.1cm}
\end{figure}

\noindent\textbf{Where the speedup comes from.}
Fig.~\ref{fig:loop_breakdown} summarizes the total decode time and the output tokens by the number of drafting stages $S$ over the math benchmarks. DLoop reduces the total decode time for all four draft models by 4.9--24.3\%. Among the generated tokens, 34--71\% are produced in loops with $S>1$, where multiple drafting stages are covered by a single verification rather than one verification per drafting stage. Table~\ref{tab:dloop-loops-by-S-math} shows what one verification yields by $S$. The time per loop increases with $S$, as more drafting stages are performed before each verification. However, the relative increase in $\tau$ with $S$ is larger than that in the time per loop, resulting in a monotonic decrease in the time per output token from 4.60--6.42\,ms at $S=1$ to 0.68--2.97\,ms at $S\geq5$. Overall, DLoop reduces the number of verifications by 7.6--34.1\%. Since each verification requires a forward pass of the target model and dominates the cost of a loop, the reduction in verifications outweighs the additional drafting stages. DLoop therefore obtains its speedup by spending additional draft-model forward passes to reduce the number of target-model forward passes.

\begin{table}[t]
\centering
\caption{\textbf{Time and output tokens per loop by the number of drafting stages.}
Results on the math benchmarks with Qwen3-4B.
Loops is the number of loops (i.e., verifications), ms/L the average time per loop, $\tau/L$ the acceptance length per loop, and ms/tok the average time per output token.
No loop denotes the draft model without DLoop, and All $S$ aggregates all loops of DLoop.}
\label{tab:dloop-loops-by-S-math}
\setlength{\tabcolsep}{3pt}
\resizebox{\textwidth}{!}{\small
\begin{tabular}{@{}l rrrr rrrr rrrr rrrr@{}}
\toprule
 & \multicolumn{4}{c}{\textbf{EAGLE-3}} & \multicolumn{4}{c}{\textbf{DFlash}} & \multicolumn{4}{c}{\textbf{Domino}} & \multicolumn{4}{c}{\textbf{DSpark}} \\
\cmidrule(lr){2-5}\cmidrule(lr){6-9}\cmidrule(lr){10-13}\cmidrule(lr){14-17}
 & Loops & ms/L & \multicolumn{1}{c}{$\tau$L} & ms/tok & Loops & ms/L & \multicolumn{1}{c}{$\tau$/L} & ms/tok & Loops & ms/L & \multicolumn{1}{c}{$\tau$/L} & ms/tok & Loops & ms/L & \multicolumn{1}{c}{$\tau$/L} & ms/tok \\
\midrule
No loop & 24,361 & 40.91 & 7.88 & 5.259 & 24,839 & 27.36 & 8.00 & 3.643 & 23,061 & 30.11 & 8.83 & 3.717 & 30,471 & 27.84 & 6.17 & 4.614 \\
\midrule
$S{=}1$ & 11,974 & 40.91 & 7.14 & 5.934 & 19,702 & 27.36 & 6.28 & 4.625 & 18,719 & 30.11 & 7.06 & 4.598 & 12,481 & 27.84 & 4.65 & 6.423 \\
$S{=}2$ & 2,828 & 54.10 & 14.23 & 3.763 & 2,068 & 32.53 & 20.60 & 1.576 & 1,977 & 37.69 & 20.96 & 1.798 & 3,091 & 32.33 & 10.09 & 3.238 \\
$S{=}3$ & 1,361 & 67.29 & 20.88 & 3.239 & 465 & 37.70 & 31.91 & 1.180 & 435 & 45.27 & 33.01 & 1.425 & 1,585 & 36.81 & 14.69 & 2.510 \\
$S{=}4$ & 709 & 80.48 & 27.33 & 2.977 & 148 & 42.87 & 44.34 & 0.958 & 116 & 52.86 & 50.58 & 1.203 & 982 & 41.30 & 19.27 & 2.183 \\
$S{\ge}5$ & 473 & 99.44 & 32.80 & 2.968 & 114 & 55.70 & 73.01 & 0.681 & 61 & 67.90 & 66.91 & 1.060 & 1,944 & 53.13 & 30.80 & 1.735 \\
\addlinespace[1.0pt]
\cmidrule(lr){1-17}
\textbf{All $S$} & \textbf{17,345} & \textbf{48.34} & \textbf{12.04} & \textbf{4.502} & \textbf{22,497} & \textbf{28.30} & \textbf{9.48} & \textbf{3.311} & \textbf{21,308} & \textbf{31.35} & \textbf{10.44} & \textbf{3.535} & \textbf{20,083} & \textbf{32.35} & \textbf{11.02} & \textbf{3.493} \\
\bottomrule
\end{tabular}}
\end{table}

\begin{table}[t]
\centering
\caption{\textbf{Results with tree verification.}
Complementary to tree verification, DLoop further raises the average acceptance length ($\tau$) and speedup (spd) of every draft model.
For each draft model, the first row is the publicly released draft model.
``$+$~Tree'' applies tree verification to it.
``$+$~Tree~$+$~DLoop'' further applies DLoop.
Bold marks the best result for each draft model.
Per-benchmark results are given in Table~\ref{tab:full} in the appendix.}
\label{tab:main_tree}
\setlength{\tabcolsep}{2.5pt}
\renewcommand{\arraystretch}{1.0}
\resizebox{\textwidth}{!}{\footnotesize
\begin{tabular}{@{}l*{4}{r}c*{4}{r}@{}}
\toprule
 & \multicolumn{4}{c}{\textbf{Qwen3-4B}} & & \multicolumn{4}{c}{\textbf{Qwen3-8B}} \\
\cmidrule(lr){2-5} \cmidrule(l){7-10}
\multirow{2}{*}{Method} & \multicolumn{1}{c}{\textsc{Math}} & \multicolumn{1}{c}{\textsc{Code}} & \multicolumn{1}{c}{\textsc{Chat}} & \multicolumn{1}{c}{\textit{Avg.}} & & \multicolumn{1}{c}{\textsc{Math}} & \multicolumn{1}{c}{\textsc{Code}} & \multicolumn{1}{c}{\textsc{Chat}} & \multicolumn{1}{c}{\textit{Avg.}} \\
\cmidrule(lr){2-2} \cmidrule(lr){3-3} \cmidrule(lr){4-4} \cmidrule(lr){5-5} \cmidrule(lr){7-7} \cmidrule(lr){8-8} \cmidrule(lr){9-9} \cmidrule(lr){10-10}
 & $\tau$\,/\,spd & $\tau$\,/\,spd & $\tau$\,/\,spd & $\tau$\,/\,spd & & $\tau$\,/\,spd & $\tau$\,/\,spd & $\tau$\,/\,spd & $\tau$\,/\,spd \\
\midrule
DFlash~\citeyear{chen2026dflash}  & 7.19\,/\,5.02$\times$ & 6.59\,/\,4.65$\times$ & 3.74\,/\,2.27$\times$ & 6.10\,/\,4.19$\times$ & & 7.22\,/\,5.13$\times$ & 6.56\,/\,4.45$\times$ & 3.66\,/\,2.22$\times$ & 6.08\,/\,4.15$\times$ \\
\;\;$+$ DDTree~\citeyear{ringel2026ddtree}                      & 9.66\,/\,6.73$\times$ & 9.23\,/\,6.33$\times$ & 5.67\,/\,3.56$\times$ & 8.50\,/\,5.79$\times$ & & 9.66\,/\,6.82$\times$ & 9.25\,/\,6.44$\times$ & 5.56\,/\,3.57$\times$ & 8.48\,/\,5.87$\times$ \\
\;\;$+$ DDTree $+$ \textbf{DLoop}   & \textbf{13.27}\,/\,\textbf{8.11}$\times$ & \textbf{10.20}\,/\,\textbf{6.62}$\times$ & \textbf{6.27}\,/\,\textbf{3.81}$\times$ & \textbf{10.37}\,/\,\textbf{6.48}$\times$ & & \textbf{12.97}\,/\,\textbf{8.35}$\times$ & \textbf{10.18}\,/\,\textbf{6.78}$\times$ & \textbf{6.12}\,/\,\textbf{3.79}$\times$ & \textbf{10.22}\,/\,\textbf{6.62}$\times$ \\
\midrule
Domino~\citeyear{huang2026domino} & 8.85\,/\,5.63$\times$ & 6.99\,/\,4.39$\times$ & 4.56\,/\,2.48$\times$ & 7.08\,/\,4.38$\times$ & & 9.04\,/\,5.75$\times$ & 7.18\,/\,4.49$\times$ & 4.54\,/\,2.51$\times$ & 7.22\,/\,4.47$\times$ \\
\;\;$+$ DARTree~\citeyear{li2026dartree}                      & 11.78\,/\,6.15$\times$ & 9.83\,/\,5.26$\times$ & 6.58\,/\,3.14$\times$ & 9.75\,/\,5.06$\times$ & & 11.86\,/\,6.31$\times$ & 10.06\,/\,5.29$\times$ & 6.59\,/\,3.10$\times$ & 9.87\,/\,5.13$\times$ \\
\;\;$+$ DARTree $+$ \textbf{DLoop}   & \textbf{14.22}\,/\,\textbf{6.83}$\times$ & \textbf{10.81}\,/\,\textbf{5.43}$\times$ & \textbf{7.12}\,/\,\textbf{3.18}$\times$ & \textbf{11.17}\,/\,\textbf{5.40}$\times$ & & \textbf{15.08}\,/\,\textbf{6.99}$\times$ & \textbf{11.38}\,/\,\textbf{5.59}$\times$ & \textbf{7.21}\,/\,\textbf{3.20}$\times$ & \textbf{11.73}\,/\,\textbf{5.52}$\times$ \\
\midrule
DSpark~\citeyear{cheng2026dspark} & 6.26\,/\,4.36$\times$ & 5.61\,/\,3.96$\times$ & 4.09\,/\,2.62$\times$ & 5.47\,/\,3.77$\times$ & & 6.32\,/\,4.46$\times$ & 5.81\,/\,3.96$\times$ & 4.13\,/\,2.61$\times$ & 5.58\,/\,3.81$\times$ \\
\;\;$+$ DARTree~\citeyear{li2026dartree}                      & 6.94\,/\,4.61$\times$ & 6.78\,/\,4.47$\times$ & 5.23\,/\,3.25$\times$ & 6.45\,/\,4.22$\times$ & & 7.08\,/\,4.58$\times$ & 6.93\,/\,4.40$\times$ & 5.42\,/\,3.35$\times$ & 6.61\,/\,4.21$\times$ \\
\;\;$+$ DARTree $+$ \textbf{DLoop}   & \textbf{12.60}\,/\,\textbf{6.10}$\times$ & \textbf{10.08}\,/\,\textbf{5.40}$\times$ & \textbf{6.52}\,/\,\textbf{3.35}$\times$ & \textbf{10.14}\,/\,\textbf{5.15}$\times$ & & \textbf{12.31}\,/\,\textbf{6.36}$\times$ & \textbf{10.16}\,/\,\textbf{5.42}$\times$ & \textbf{6.72}\,/\,\textbf{3.45}$\times$ & \textbf{10.11}\,/\,\textbf{5.28}$\times$ \\
\bottomrule
\end{tabular}}
\end{table}

\noindent\textbf{DLoop with tree verification. }
\label{sec:tree_results}
Table~\ref{tab:main_tree} shows that DLoop further raises the acceptance length and the speedup when applied on top of tree verification.
Tree verification broadens the draft with branches that hold alternative tokens at the same position, whereas DLoop lengthens it with additional drafting stages.
On a tree, DLoop extends the highest-scoring path out of tree branches.
The gate function scores this path as in Section~\ref{sec:loop}.
The draft tokens of each additional drafting stage form another tree rooted at the end of this path for DDTree and a single chain for DARTree.
The target model verifies the tree and the added draft tokens together in one forward pass.
Over tree verification alone, DLoop raises the average speedup of DFlash with DDTree by 12 to 13 percent.
With DARTree, the gain is 7 to 8 percent for Domino and 22 to 25 percent for DSpark.
Since EAGLE-3 uses its own tree verification, its results in Table~\ref{tab:main} already show this gain.

\section{Conclusion}
We introduced DLoop, a looped form of speculative decoding in which a single verification covers multiple drafting stages. The confidence-gated drafting loop repeats the drafting stage while the draft model remains confident. Loop-aware training keeps the draft model reliable in the additional drafting stages by exposing it to its own hidden states for unverified draft tokens. DLoop leaves the architecture of the draft model and the acceptance rule unchanged. It changes only how many drafting stages precede a verification. Across EAGLE-3, DFlash, Domino, DSpark, and multi-token prediction modules, DLoop improves the wall-clock speedup by 5 to 41 percent while preserving lossless decoding. It obtains this speedup by spending additional draft-model forward passes to reduce the number of target-model forward passes. Adaptive draft length methods raise the speedup only for autoregressive draft models, whereas DLoop raises it for parallel draft models as well. The ablation shows that loop-aware training prepares the draft model for the loop rather than improving each drafting stage on its own. Applied on top of tree verification, DLoop further raises the acceptance length and the speedup. These results show that the number of drafting stages before a verification need not be fixed by the schedule and can be decided during decoding.

\clearpage

\bibliography{iclr2027_conference}
\bibliographystyle{iclr2027_conference}

\clearpage
\appendix
\section{Appendix}
\setcounter{table}{0}
\setcounter{figure}{0}
\renewcommand{\thetable}{A\arabic{table}}
\renewcommand{\thefigure}{A\arabic{figure}}
\renewcommand{\theHtable}{A\arabic{table}}
\renewcommand{\theHfigure}{A\arabic{figure}}

\begin{list}{\labelitemi}{\settowidth{\labelwidth}{\labelitemi}\setlength{\leftmargin}{\labelwidth}\addtolength{\leftmargin}{\labelsep}\setlength{\topsep}{0pt}\setlength{\partopsep}{0pt}\setlength{\itemsep}{1pt}\setlength{\parsep}{0pt}}
\item Appendix~\ref{app:protocol} describes the evaluation protocol, including the benchmark sources, the sample counts, and the prompt templates.
\item Appendix~\ref{app:per_benchmark} reports the results on each benchmark with greedy decoding and with sampling at temperature $T=1$.
\item Appendix~\ref{app:serving} evaluates the decoding throughput of DLoop in SGLang under concurrent serving.
\item Appendix~\ref{app:calls} analyzes how DLoop changes the total numbers of drafting stages and verifications.
\item Appendix~\ref{app:loop_cost_by_S} measures the time and the peak GPU memory of one loop by the number of drafting stages.
\item Appendix~\ref{app:examples} presents representative generations of the confidence-gated drafting loop.
\item Appendix~\ref{app:design_choices} compares design choices for the gate threshold, the gate function, and loop-aware training.
\item Appendix~\ref{app:adaptive_baselines} details how we apply the adaptive draft length methods to EAGLE-3 and DSpark.
\end{list}

\subsection{Evaluation protocol}
\label{app:protocol}
The eight benchmarks and their sample counts follow the official DFlash evaluation code. Table~\ref{tab:protocol_samples} lists the source and the number of evaluated samples for each benchmark. Each subset of samples consists of the first samples upto the number of samples after shuffling the benchmark with seed 0. Table~\ref{tab:protocol_prompts} lists the prompt templates of the DFlash evaluation code. MBPP and MT-Bench use the text of each sample without a template. We place each prompt in a user message and apply the chat template of the target model. We do not add a system prompt and disable thinking mode. The instruction at the start of the LiveCodeBench (LCB) template is therefore part of the user message. Every result on the eight benchmarks follows this protocol.

\begin{table}[h]
\centering\small
\caption{\textbf{Number of evaluated samples per benchmark.} Size is the number of samples in the source dataset on Hugging Face.}
\label{tab:protocol_samples}
\resizebox{\columnwidth}{!}{%
\begin{tabular}{llrr}
\toprule
Benchmark & Source & Size & Samples \\
\midrule
\multicolumn{4}{l}{\textsc{Math}} \\
GSM8K~\citep{cobbe2021gsm8k}     & \texttt{openai/gsm8k}                        & 1,319  & 128 \\
MATH-500~\citep{lightman2023math500}  & \texttt{HuggingFaceH4/MATH-500}              & 500    & 128 \\
AIME25    & \texttt{MathArena/aime\_2025}                & 30     & 30  \\
\midrule
\multicolumn{4}{l}{\textsc{Code}} \\
HumanEval~\citep{chen2021humaneval} & \texttt{openai/openai\_humaneval}            & 164    & 164 \\
MBPP~\citep{austin2021mbpp}      & \texttt{google-research-datasets/mbpp}       & 257    & 128 \\
LCB~\citep{jain2025livecodebench}       & \texttt{livecodebench/code\_generation\_lite} & 1,055  & 128 \\
\midrule
\multicolumn{4}{l}{\textsc{Chat}} \\
MT-Bench~\citep{zheng2023mtbench}  & \texttt{HuggingFaceH4/mt\_bench\_prompts}    & 80     & 80  \\
Alpaca~\citep{taori2023alpaca}    & \texttt{tatsu-lab/alpaca}                    & 52,002 & 128 \\
\bottomrule
\end{tabular}%
}
\end{table}

\subsection{Per-benchmark results}
\label{app:per_benchmark}
Table~\ref{tab:full} reports the acceptance length and the speedup of every setting in Tables~\ref{tab:main}, \ref{tab:dloop_ablation}, and \ref{tab:main_tree} on each benchmark. DLoop raises the acceptance length on every benchmark for every draft model and target model. The gain is largest on math and smallest on chat. The full acceptance ratio in Fig.~\ref{fig:full_acceptance} follows the same order. Table~\ref{tab:full_t1} reports the same settings with sampling at temperature $T=1$. At $T=1$, DLoop also raises the average acceptance length.

\subsection{Serving throughput under concurrency.}
\label{app:serving}
Table~\ref{tab:sglang_concurrency} evaluates DLoop with both DFlash and DSpark in SGLang under concurrent serving with Qwen3-8B as the target model. DLoop improves throughput substantially at low concurrency. On MATH-500 at $c=1$, the speedup increases from 5.81$\times$ to 6.23$\times$ with DFlash and from 4.45$\times$ to 4.94$\times$ with DSpark, with similar gains on HumanEval. The gain generally becomes smaller as concurrency increases, but DLoop remains beneficial through $c=8$ with DFlash and through $c=16$ with DSpark on both benchmarks. At $c=32$, DFlash+DLoop slightly underperforms DFlash, whereas DSpark+DLoop continues to improve throughput on MATH-500 and remains within 1\% of DSpark on HumanEval. This attenuation is consistent with the known interaction between speculative decoding and batching, in which additional speculative work consumes increasingly scarce batch capacity as the target model becomes saturated, reducing the optimal speculation depth at high load~\citep{su2023synergy,liu2024turbospec,cheng2026dspark}. DLoop further chooses the number of drafting stages independently for each sample, yielding ragged draft-side execution whose efficient batching requires either padding, which wastes computation, or dynamic regrouping, which introduces additional scheduling overhead. Such sample-dependent execution therefore leaves room for serving and kernel optimizations that better support adaptive loop depths. Overall, DLoop retains substantial throughput gains at low-to-moderate concurrency, while its advantage diminishes as serving becomes increasingly batch-bound.

\begin{table}[!ht]
\caption{\textbf{SGLang decoding throughput with DFlash, DSpark, and DLoop under different concurrency levels.}
We report throughput in tokens per second (tok/s), followed by speedup over the autoregressive (AR) baseline. \textbf{Bold} indicates the better result within each model pair.}
\label{tab:sglang_concurrency}
\centering
\resizebox{\columnwidth}{!}{
\begin{tabular}{lcccccc}
\toprule
Method & $c=1$ & $c=2$ & $c=4$ & $c=8$ & $c=16$ & $c=32$ \\
\midrule
\multicolumn{7}{l}{\textbf{MATH-500}} \\
AR & 156 / 1.00$\times$ & 303 / 1.00$\times$ & 593 / 1.00$\times$ & 1,138 / 1.00$\times$ & 2,132 / 1.00$\times$ & 3,763 / 1.00$\times$ \\
DFlash & 907 / 5.81$\times$ & 1,750 / 5.78$\times$ & 3,252 / 5.48$\times$ & 6,019 / 5.29$\times$ & \textbf{9,484} / \textbf{4.45$\times$} & \textbf{11,810} / \textbf{3.14$\times$} \\
\quad + \textbf{DLoop} & \textbf{972} / \textbf{6.23$\times$} & \textbf{1,833} / \textbf{6.05$\times$} & \textbf{3,362} / \textbf{5.67$\times$} & \textbf{6,134} / \textbf{5.39$\times$} & 9,189 / 4.31$\times$ & 11,139 / 2.96$\times$ \\
DSpark & 694 / 4.45$\times$ & 1,244 / 4.11$\times$ & 2,517 / 4.24$\times$ & 4,538 / 3.99$\times$ & 8,117 / 3.81$\times$ & 12,265 / 3.26$\times$ \\
\quad + \textbf{DLoop} & \textbf{770} / \textbf{4.94$\times$} & \textbf{1,480} / \textbf{4.88$\times$} & \textbf{2,692} / \textbf{4.54$\times$} & \textbf{4,874} / \textbf{4.28$\times$} & \textbf{8,266} / \textbf{3.88$\times$} & \textbf{12,820} / \textbf{3.41}$\times$ \\
\midrule
\multicolumn{7}{l}{\textbf{HumanEval}} \\
AR & 156 / 1.00$\times$ & 301 / 1.00$\times$ & 595 / 1.00$\times$ & 1,149 / 1.00$\times$ & 2,154 / 1.00$\times$ & 3,762 / 1.00$\times$ \\
DFlash & 717 / 4.60$\times$ & 1,357 / 4.51$\times$ & 2,552 / 4.29$\times$ & 4,673 / 4.07$\times$ & 7,489 / 3.48$\times$ & \textbf{9,451} / \textbf{2.51$\times$} \\
\quad + \textbf{DLoop} & \textbf{779} / \textbf{4.98$\times$} & \textbf{1,454} / \textbf{4.83$\times$} & \textbf{2,648} / \textbf{4.45$\times$} & \textbf{4,791} / \textbf{4.17$\times$} & \textbf{7,668} / \textbf{3.56$\times$} & 9,104 / 2.42$\times$ \\
DSpark & 641 / 4.11$\times$ & 1,228 / 4.07$\times$ & 2,243 / 3.77$\times$ & 4,232 / 3.68$\times$ & 7,234 / 3.36$\times$ & \textbf{10,926} / \textbf{2.90$\times$} \\
\quad + \textbf{DLoop} & \textbf{714} / \textbf{4.58$\times$} & \textbf{1,326} / \textbf{4.41$\times$} & \textbf{2,451} / \textbf{4.12$\times$} & \textbf{4,317} / \textbf{3.76$\times$} & \textbf{7,297} / \textbf{3.39$\times$} & 10,825 / 2.88$\times$ \\
\bottomrule
\end{tabular}
}
\end{table}

\subsection{Change in the number of drafting stages and verifications}
\label{app:calls}
Table~\ref{tab:dloop-calls} shows how DLoop changes the total numbers of drafting stages and verifications over the eight benchmarks.
Without tree verification, DLoop increases the number of drafting stages by 3.5--24.5\% and reduces the number of verifications by 4.3--22.7\% across the four draft models.
On top of tree verification, DLoop increases the number of drafting stages by 4.4--37.9\% and reduces the number of verifications by 4.6--26.9\%.
In both settings, the gains in acceptance length and speedup rank the draft models in the same order as the reduction in verifications.
The reduction is largest for DSpark and smallest for Domino.
For DSpark and for DFlash with tree verification, the added drafting stages outnumber the removed verifications.
The speedup still increases by 10.2--27.6\% in these settings.
Table~\ref{tab:loop_cost_by_S} shows that each additional drafting stage takes less time than one verification for every draft model and every $S$.
The reduction in verifications therefore outweighs the additional drafting stages.

\begin{table}[!ht]
\centering\small
\setlength{\tabcolsep}{4pt}
\caption{\textbf{Number of drafting stages and verifications.} The baseline performs one drafting stage per verification. DLoop trades additional drafting stages for fewer verifications. The counts are measured on Qwen3-4B over the eight benchmarks, and the baseline is the loop-aware tuned draft model. The changes in acceptance length ($\tau$) and speedup (spd) are computed from the corresponding rows of Table~\ref{tab:main}.}
\label{tab:dloop-calls}
\begin{tabular}{l rr rr rrrr}
\toprule
& \multicolumn{2}{c}{\textbf{Baseline}} & \multicolumn{2}{c}{\textbf{+ DLoop}} & \multicolumn{4}{c}{\textbf{Change}} \\
\cmidrule(lr){2-3}\cmidrule(lr){4-5}\cmidrule(lr){6-9}
Setup & Drafting & Verification & Drafting & Verification & \multicolumn{1}{c}{$\tau$} & Speedup & Drafting & Verification \\
\midrule
EAGLE-3        & 70,311 & 70,311 & 78,119  & 57,594 & $+$37.8\% & $+$11.9\% & $+$11.1\% & $-$18.1\% \\
\midrule
DFlash         & 87,172 & 87,172 & 90,601  & 82,611 & $+$13.0\% & $+$10.4\% & $+$3.9\%  & $-$5.2\% \\
\quad + Tree   & 58,924 & 58,924 & 65,162  & 53,721 & $+$15.5\% & $+$10.2\% & $+$10.6\% & $-$8.8\% \\
\midrule
Domino         & 78,415 & 78,415 & 81,192  & 75,023 & $+$11.9\% & $+$4.6\%  & $+$3.5\%  & $-$4.3\% \\
\quad + Tree   & 51,689 & 51,689 & 53,989  & 49,314 & $+$8.1\%  & $+$7.1\%  & $+$4.4\%  & $-$4.6\% \\
\midrule
DSpark         & 91,851 & 91,851 & 114,391 & 70,978 & $+$58.8\% & $+$27.6\% & $+$24.5\% & $-$22.7\% \\
\quad + Tree   & 75,753 & 75,753 & 104,460 & 55,349 & $+$47.8\% & $+$25.3\% & $+$37.9\% & $-$26.9\% \\
\bottomrule
\end{tabular}
\end{table}

\subsection{Cost of a loop by the number of drafting stages}
\label{app:loop_cost_by_S}
Table~\ref{tab:loop_cost_by_S} reports the time and the peak GPU memory of one loop for $S$ from 1 to 4.
At $S=1$, the loop reduces to standard speculative decoding.
For every draft model, the drafting time increases almost linearly with $S$.
The verification time remains within 21.9--25.0\,ms, although the verification covers more draft tokens as $S$ grows.
One drafting stage costs less than one verification, even for EAGLE-3.
The peak GPU memory increases by less than 0.1\,GB from $S=1$ to $S=4$.

\begin{table}[!ht]
\centering
\caption{\textbf{Latency and peak GPU memory by the number of drafting stages.}
Each loop runs a fixed number $S$ of drafting stages and one verification.
Measured with Qwen3-4B~\citep{yang2025qwen3} on a single H100 GPU in bf16 with a 1{,}024-token prefix.
Draft is the total time of the $S$ drafting stages in ms.
Verify is the time of one verification over all draft tokens in ms.
The gate function takes at most 0.33\,ms per loop and is excluded.}
\label{tab:loop_cost_by_S}
\setlength{\tabcolsep}{3pt}
\small
\begin{tabular}{@{}c rrr rrr rrr rrr@{}}
\toprule
 & \multicolumn{3}{c}{\textbf{EAGLE-3}} & \multicolumn{3}{c}{\textbf{DFlash}} & \multicolumn{3}{c}{\textbf{Domino}} & \multicolumn{3}{c}{\textbf{DSpark}} \\
\cmidrule(lr){2-4}\cmidrule(lr){5-7}\cmidrule(lr){8-10}\cmidrule(l){11-13}
$S$ & Draft & Verify & Mem. & Draft & Verify & Mem. & Draft & Verify & Mem. & Draft & Verify & Mem. \\
\midrule
1 & 12.10 & 23.09 & 9.08 & 3.66  & 21.99 & 8.91 & 6.00  & 22.75 & 9.01 & 3.90  & 22.45 & 9.06 \\
2 & 23.13 & 22.63 & 9.10 & 7.23  & 22.46 & 8.92 & 11.70 & 22.52 & 9.02 & 7.58  & 23.24 & 9.06 \\
3 & 34.72 & 22.78 & 9.13 & 10.83 & 23.30 & 8.92 & 17.69 & 23.18 & 9.02 & 12.41 & 23.32 & 9.06 \\
4 & 47.32 & 23.37 & 9.16 & 13.96 & 23.56 & 8.93 & 23.81 & 23.18 & 9.03 & 15.04 & 24.32 & 9.06 \\
\bottomrule
\end{tabular}
\end{table}

\subsection{Examples of the drafting loop.}
\label{app:examples}
\begin{figure}[!ht]
\centering
    \includegraphics[width=\linewidth]{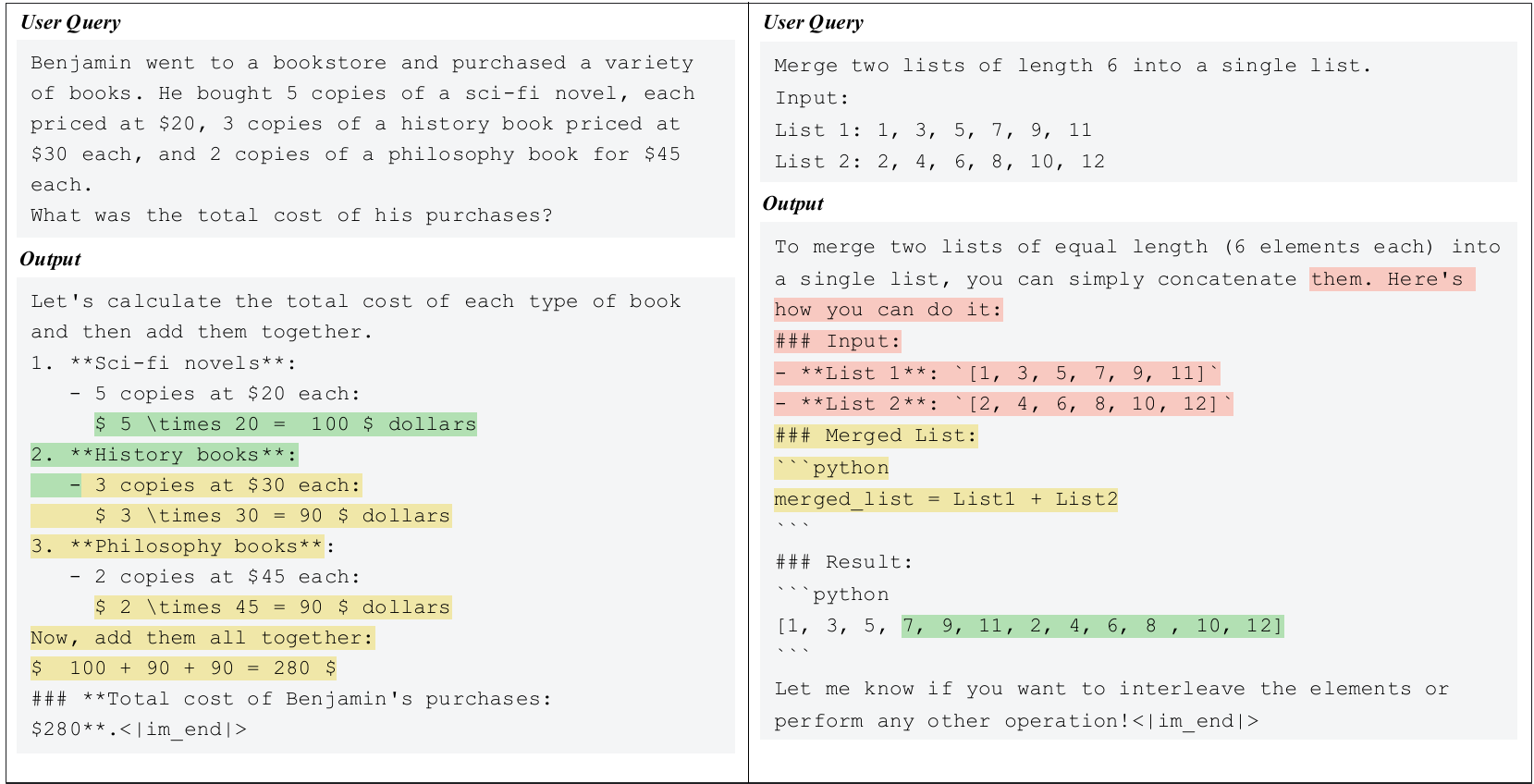}
    \caption{\textbf{Examples of the confidence-gated drafting loop.}
    We show representative generations illustrating how the number of drafting stages $S$ varies across loops.
    Highlighted spans indicate output tokens generated in loops with
    \begingroup
    \setlength{\fboxsep}{1pt}
    \protect\colorbox{loopgreen}{$S=2$},
    \protect\colorbox{loopyellow}{$S=3$}, and
    \protect\colorbox{loopred}{$S=4$}
    \endgroup
    drafting stages, respectively; unhighlighted spans correspond to loops with $S=1$.
    In each loop, DLoop performs the indicated number of drafting stages before a single target-model verification.}
    \label{fig:examples_main}
    \vspace{-0.1cm}
\end{figure}

Fig.~\ref{fig:examples_main} shows representative generations with DSpark to illustrate how the confidence-gated drafting loop behaves within an output. The number of drafting stages $S$ varies across loops. Some loops stop after the first drafting stage ($S=1$), while confident loops continue for two, three, or four drafting stages before a single verification, as indicated by the highlighted spans. The examples show that DLoop does not use a fixed number of drafting stages throughout generation, but selectively extends drafting only in loops where the gate permits it. This allows multiple consecutive blocks to be covered by one target-model verification and reduces the end-to-end decoding latency compared with the same DSpark model without the loop.

\subsection{Design Choices for Gate Function and Loop-Aware Training}
\label{app:design_choices}

\subsubsection{Gate threshold.}
Table~\ref{tab:gate_sweep} reports the average acceptance length and speedup of DLoop for each gate threshold $g$.

\begin{table}[!ht]
\centering
\caption{\textbf{Gate threshold sweep.}
Each cell reports acceptance length and speedup ($\tau$\,/\,spd$\times$) of DLoop averaged over the eight benchmarks at $T=0$.}
\label{tab:gate_sweep}
\setlength{\tabcolsep}{2.5pt}
\renewcommand{\arraystretch}{1.0}
\resizebox{\textwidth}{!}{\footnotesize
\begin{tabular}{@{}l*{10}{c}@{}}
\toprule
\multirow{2}{*}{$g$} & \multicolumn{2}{c}{\textbf{EAGLE-3}} & \multicolumn{2}{c}{\textbf{DFlash}} & \multicolumn{2}{c}{\textbf{Domino}} & \multicolumn{2}{c}{\textbf{DSpark}} & \multicolumn{2}{c}{\textbf{MTP}} \\
\cmidrule(lr){2-3} \cmidrule(lr){4-5} \cmidrule(lr){6-7} \cmidrule(lr){8-9} \cmidrule(l){10-11}
 & Q4B & Q8B & Q4B & Q8B & Q4B & Q8B & Q4B & Q8B & Q9B & G12B \\
\midrule
$-0.25$ & 9.01\,/\,3.71$\times$ & 7.59\,/\,3.44$\times$ & 7.21\,/\,4.60$\times$ & 7.39\,/\,4.66$\times$ & 7.83\,/\,4.51$\times$ & 8.19\,/\,4.74$\times$ & 8.51\,/\,4.70$\times$ & 8.54\,/\,4.74$\times$ & 5.41\,/\,2.06$\times$ & 6.35\,/\,3.58$\times$ \\
$-0.5$ & 9.63\,/\,\textbf{3.77}$\times$ & 7.87\,/\,3.46$\times$ & 7.21\,/\,\textbf{4.67}$\times$ & 7.40\,/\,\textbf{4.71}$\times$ & 7.90\,/\,\textbf{4.58}$\times$ & 8.29\,/\,\textbf{4.76}$\times$ & 8.59\,/\,\textbf{4.76}$\times$ & 9.08\,/\,\textbf{4.82}$\times$ & 6.70\,/\,\textbf{2.18}$\times$ & 7.50\,/\,\textbf{3.68}$\times$ \\
$-0.75$ & 10.12\,/\,3.75$\times$ & 8.22\,/\,\textbf{3.51}$\times$ & 7.29\,/\,4.52$\times$ & 7.50\,/\,4.49$\times$ & 8.03\,/\,4.51$\times$ & 8.52\,/\,4.69$\times$ & 9.14\,/\,4.52$\times$ & 9.40\,/\,4.69$\times$ & 7.10\,/\,2.14$\times$ & 7.82\,/\,3.65$\times$ \\
$-1.0$ & 10.77\,/\,3.55$\times$ & 8.51/3.42$\times$ & 7.35\,/\,4.48$\times$ & 7.58\,/\,4.41$\times$ & 8.15\,/\,4.48$\times$ & 8.62\,/\,4.59$\times$ & 9.33\,/\,4.22$\times$ & 9.69\,/\,4.40$\times$ & 7.46\,/\,2.10$\times$ & 8.23\,/\,3.41$\times$ \\
\bottomrule
\end{tabular}}
\end{table}

\subsubsection{Gate function.}
Table~\ref{tab:gate} compares the log-probability score used by DLoop with alternative confidence scores under the same drafting budget. The log probability gives the highest AUROC and precision for all four draft models, and also gives the highest speedup among the practical gate functions. The learned per-token head attains slightly longer acceptance length for DFlash and Domino, but this does not translate into higher speedup. The learned heads require additional parameters and a separate training stage, yet remain below the parameter-free scores in AUROC and precision. Entropy is the strongest parameter-free alternative, but is consistently below the log probability in AUROC and precision. Entropy measures uncertainty over the entire draft distribution, whereas the log probability directly measures the probability assigned to the draft tokens that will be checked by the acceptance rule. Detailed threshold-sweep results for each gate function are provided in the appendix.

\begin{table}[!ht]
\centering
\caption{\textbf{Confidence scores for the gate with Qwen3-4B.}
A block is positive if all of its draft tokens are accepted; for EAGLE-3, a block is the highest-scoring root-to-leaf path. AUROC measures ranking quality, while precision is the fraction of continued blocks that are positive, evaluated at the same continuation rate as our log-probability gate. Acceptance length ($\tau$) and speedup (spd) are averaged over eight benchmarks. Tree verification is disabled except for EAGLE-3, and Oracle continues only on positive blocks. \textbf{Bold} indicates the best non-oracle result for each metric}
\label{tab:gate}
\setlength{\tabcolsep}{3pt}
\resizebox{\textwidth}{!}{\small
\begin{tabular}{@{}l*{4}{rrrr}@{}}
\toprule
 & \multicolumn{4}{c}{\textbf{EAGLE-3}} & \multicolumn{4}{c}{\textbf{DFlash}} & \multicolumn{4}{c}{\textbf{Domino}} & \multicolumn{4}{c}{\textbf{DSpark}} \\
\cmidrule(lr){2-5} \cmidrule(lr){6-9} \cmidrule(lr){10-13} \cmidrule(l){14-17}
Gate & \multicolumn{1}{c}{AUROC} & \multicolumn{1}{c}{Prec.} & \multicolumn{1}{c}{$\tau$} & \multicolumn{1}{c}{spd} & \multicolumn{1}{c}{AUROC} & \multicolumn{1}{c}{Prec.} & \multicolumn{1}{c}{$\tau$} & \multicolumn{1}{c}{spd} & \multicolumn{1}{c}{AUROC} & \multicolumn{1}{c}{Prec.} & \multicolumn{1}{c}{$\tau$} & \multicolumn{1}{c}{spd} & \multicolumn{1}{c}{AUROC} & \multicolumn{1}{c}{Prec.} & \multicolumn{1}{c}{$\tau$} & \multicolumn{1}{c}{spd} \\
\midrule
No loop                   & --    & --    & 6.99 & 3.37 & --    & --    & 6.38 & 4.23 & --    & --    & 7.06 & 4.38 & --    & --    & 5.41 & 3.73 \\
\midrule
DSpark head               & --    & --    & --   & --   & --    & --    & --   & --   & --    & --    & --   & --   & 0.909 & 75.0 & 8.47 & 4.68 \\
Learned head (per token)  & 0.810 & 58.3 & 9.31 & 3.62 & 0.975 & 69.9 & \textbf{7.24} & 4.57 & 0.927 & 63.3 & \textbf{7.94} & 4.45 & 0.907 & 74.7 & 8.56 & 4.65 \\
Learned head (pooled)     & 0.799 & 56.6 & 8.81 & 3.58 & 0.974 & 69.1 & 7.17 & 4.57 & 0.924 & 61.0 & 7.58 & 4.43 & 0.909 & 75.3 & 8.13 & 4.67 \\
Entropy                   & 0.838 & 67.1 & 9.25 & 3.71 & 0.979 & 72.9 & 7.21 & 4.61 & 0.960 & 75.3 & 7.84 & 4.52 & 0.922 & 77.3 & 8.51 & 4.74 \\
\textbf{Log prob.} (ours) & \textbf{0.852} & \textbf{69.6} & \textbf{9.63} & \textbf{3.77} & \textbf{0.983} & \textbf{75.7} & 7.21 & \textbf{4.67} & \textbf{0.963} & \textbf{76.8} & 7.90 & \textbf{4.58} & \textbf{0.924} & \textbf{77.6} & \textbf{8.59} & \textbf{4.76} \\
\midrule
Oracle                    & --    & 100.0 & 11.23 & 4.15 & --    & 100.0 & 7.35 & 4.70 & --    & 100.0 & 8.37 & 4.67 & --    & 100.0 & 9.50 & 5.32 \\
\bottomrule
\end{tabular}}
\end{table}

\subsubsection{Loop-aware training.}
Table~\ref{tab:kmax_midce} ablates depth exposure ($N$) and intermediate supervision (mid-CE) in loop-aware training. Enabling loop-aware training with $N=1$ improves the acceptance length and speedup for all three draft models, with the largest gain for DSpark (5.75 to 8.13 in $\tau$). Increasing $N$ to 3 gives further gains for DFlash and DSpark and comparable results for Domino, whereas $N=5$ gives no additional improvement despite higher training time and memory. Adding mid-CE consistently lowers the acceptance length and speedup while increasing the training cost, and runs out of memory for DSpark with $N=5$. We therefore adopt $N=3$ without mid-CE.

\begin{table}[!ht]
\caption{\textbf{Loop-aware training ablation on depth exposure and intermediate supervision.}
We report the 8-benchmark average acceptance length ($\tau$), decoding speedup, training time per iteration (Tr. s/it), and peak GPU memory (Tr. GB) with Qwen3-4B.
$N=0$ disables loop-aware training, and mid-CE applies supervision to intermediate drafting stages.
\textbf{Bold} indicates the adopted setting.}
\label{tab:kmax_midce}
\begin{center}
\setlength{\tabcolsep}{4pt}
\resizebox{\linewidth}{!}{%
\begin{tabular}{cc cccc cccc cccc}
\toprule
\multirow{2}{*}{$N$} & \multirow{2}{*}{mid-CE}
& \multicolumn{4}{c}{\textbf{DFlash}} & \multicolumn{4}{c}{\textbf{Domino}} & \multicolumn{4}{c}{\textbf{DSpark}}\\
\cmidrule(lr){3-6}\cmidrule(lr){7-10}\cmidrule(lr){11-14}
 & & $\tau$ & speedup & Tr. s/it & Tr. GB
   & $\tau$ & speedup & Tr. s/it & Tr. GB
   & $\tau$ & speedup & Tr. s/it & Tr. GB \\
\midrule
0 &  & 6.33 & 4.21$\times$ & 0.136 & 22.5 & 7.18 & 4.33$\times$ & 0.161 & 24.4 & 5.75 & 3.77$\times$ & 0.178 & 32.8 \\
\midrule
1 &  & 7.04 & 4.52$\times$ & 0.223 & 27.6 & 7.92 & 4.57$\times$ & 0.251 & 34.3 & 8.13 & 4.61$\times$ & 0.292 & 44.3 \\
\midrule
\textbf{3} &  & \textbf{7.21} & \textbf{4.67}$\times$ & \textbf{0.299} & \textbf{36.2} & \textbf{7.90} & \textbf{4.58}$\times$ & \textbf{0.324} & \textbf{42.9} & \textbf{8.59} & \textbf{4.76}$\times$ & \textbf{0.365} & \textbf{51.6} \\
3 & \checkmark & 7.07 & 4.53$\times$ & 0.321 & 38.7 & 7.87 & 4.54$\times$ & 0.349 & 47.6 & 8.42 & 4.64$\times$ & 0.440 & 65.9 \\
\midrule
5 &  & 7.13 & 4.55$\times$ & 0.361 & 45.0 & 7.90 & 4.56$\times$ & 0.392 & 52.1 & 8.52 & 4.71$\times$ & 0.427 & 59.7 \\
5 & \checkmark & 7.02 & 4.53$\times$ & 0.394 & 50.0 & 7.89 & 4.55$\times$ & 0.436 & 61.6 & \multicolumn{4}{c}{\emph{out of memory}} \\
\bottomrule
\end{tabular}%
}
\end{center}
\end{table}

\subsection{Details of the Comparison with Adaptive Draft Length}
\label{app:adaptive_baselines}

\subsubsection{Decision rules.}
BiLD, SVIP, AdaEDL, and LLMCad stop drafting once a statistic of the draft distribution crosses a threshold. BiLD uses the top-1 probability, whereas SVIP and AdaEDL use the entropy. LLMCad uses the confidence accumulated over the draft tokens. AdaEDL and LLMCad also adjust the threshold after each verification according to how many draft tokens were accepted. DDD applies a similar rule to the draft tree of EAGLE-2, checking the summed probability of the sequences in the beam at preset depths. SpecDec++ and DISCO instead train a head that decides after each draft token whether to stop. AdaEAGLE trains a predictor that sets the draft length once, before the drafting stage begins. PACER drafts in blocks of $b$ tokens ($b$ is set to each draft model's draft length) and trains a pre-verification layer. It drafts another block while the mean acceptance that this layer predicts for the latest block remains above a threshold. The confidence scheduler of DSpark chooses how many draft tokens of the full block to verify, using a confidence head trained together with DSpark. The scheduler therefore does not apply to EAGLE-3.

\subsubsection{Application to EAGLE-3.}
BiLD, SVIP, AdaEDL, LLMCad, SpecDec++, DISCO, and PACER were proposed for a separate small autoregressive LLM as the draft model. EAGLE-3 instead expands its draft tree by one depth per forward pass. We evaluate each rule after every depth on the highest-scoring path of the tree and stop the expansion once the rule stops drafting. PACER runs its pre-verification layer on the same path after every $b$ draft tokens. DDD checks the beam at the preset depths of its paper. We run each rule with the default maximum depth of EAGLE-3 and with the maximum draft length of its paper. We port the fallback policy of BiLD and omit its rollback policy. The rollback policy replaces lossless verification with a rule that can change the output of the target model.

\subsubsection{Application to DSpark.}
The parallel draft model DSpark computes all $k$ draft tokens of a drafting stage in a single forward pass. Every method except PACER and DDD evaluates its rule on the full block. Only the draft tokens before the position where the rule stops go to verification. For AdaEAGLE, the predicted length sets the number of verified draft tokens. Discarding the remaining draft tokens shortens only the verification. With Qwen3-4B on a single H100 GPU at batch size 1, a verification takes $22.2 \pm 0.3$\,ms for any number of draft tokens from 2 to 16. When a rule does not stop within the block, we also let it perform another drafting stage, up to the maximum draft length of its paper. PACER and DDD instead decide at the end of each drafting stage whether another drafting stage follows. All additional drafting stages use the publicly released weights of DSpark and take the input of Eq.~\ref{eq:loop_input}. Since DSpark drafts a single chain, the criterion of DDD becomes the sum of the log probabilities of all draft tokens after the anchor token. It differs from our gate function only in accumulating over drafting stages.

\subsubsection{Training and tuning.}
We train the decision heads of SpecDec++, DISCO, AdaEAGLE, and PACER for one epoch on the same data as loop-aware training. Their architecture, inputs, labels, and loss follow the original papers. We keep the other constants of each rule from its paper and sweep its threshold, together with $b$ for PACER. Table~\ref{tab:adaptive} reports, for each method, the setting with the highest average speedup.

\newsavebox{\protocolpromptbox}
\newlength{\protocolpromptwd}
\begin{table}[p]
\caption{\textbf{Prompt templates of the DFlash evaluation code.} Braces mark the fields filled from each sample. Long lines are wrapped for display.}
\label{tab:protocol_prompts}
\footnotesize
\settowidth{\protocolpromptwd}{\ttfamily Write a solution to the following problem and make sure that it passes the tests:}
\begin{lrbox}{\protocolpromptbox}
\begin{minipage}{\protocolpromptwd}
\textbf{GSM8K, MATH-500, AIME25}
\begin{verbatim}
{problem}
Please reason step by step, and put your final answer within \boxed{}.
\end{verbatim}

\textbf{HumanEval}
\begin{verbatim}
Write a solution to the following problem and make sure that it passes the tests:
```python
{prompt}
```
\end{verbatim}

\textbf{LCB, problems with starter code}
\begin{verbatim}
You are an expert Python programmer. You will be given a question (problem
specification) and will generate a correct Python program that matches the
specification and passes all tests. You will NOT return anything except for the
program

### Question:
{question_content}

### Format: Use the following code structure:
```python
{starter_code}
```

### Answer: (use the provided format with backticks)
\end{verbatim}

\textbf{LCB, problems without starter code}
\begin{verbatim}
### Format: Write your code in the following format:
```python
# YOUR CODE HERE
```
\end{verbatim}

\textbf{Alpaca}
\begin{verbatim}
{instruction}

Input:
{input}
\end{verbatim}
\end{minipage}
\end{lrbox}
\rule{\linewidth}{0.08em}

\resizebox{\linewidth}{!}{\usebox{\protocolpromptbox}}

\rule{\linewidth}{0.08em}
\end{table}

\begin{table}[p]
\centering
\caption{\textbf{Acceptance length ($\tau$) and speedup (spd) on each benchmark.}
Each cell reports $\tau$\,/\,spd$\times$. Avg. is the mean over the eight benchmarks. For each draft model, the first row is the publicly released draft model. ``$+$~Loop'' enables the confidence-gated drafting loop, ``$+$~Train'' applies one epoch of loop-aware training, and ``$+$~DLoop'' applies both. ``$+$~DDTree'' and ``$+$~DARTree'' apply tree verification. Bold marks the best result in each column for each draft model.}
\label{tab:full}
\setlength{\tabcolsep}{2.5pt}
\renewcommand{\arraystretch}{1.0}
\resizebox{\textwidth}{!}{\footnotesize
\begin{tabular}{@{}l*{9}{r}@{}}
\toprule
 & \multicolumn{3}{c}{\textsc{Math}} & \multicolumn{3}{c}{\textsc{Code}} & \multicolumn{2}{c}{\textsc{Chat}} & \\
\cmidrule(lr){2-4} \cmidrule(lr){5-7} \cmidrule(lr){8-9}
Method & \multicolumn{1}{c}{GSM8K} & \multicolumn{1}{c}{MATH-500} & \multicolumn{1}{c}{AIME25} & \multicolumn{1}{c}{HumanEval} & \multicolumn{1}{c}{MBPP} & \multicolumn{1}{c}{LCB} & \multicolumn{1}{c}{MT-Bench} & \multicolumn{1}{c}{Alpaca} & \multicolumn{1}{c}{\textit{Avg.}} \\
\midrule
\multicolumn{10}{c}{\textbf{Qwen3-4B}} \\
\midrule
EAGLE-3                                  & 8.11\,/\,3.99$\times$ & 7.98\,/\,4.02$\times$ & 7.41\,/\,3.54$\times$ & 7.25\,/\,3.57$\times$ & 7.25\,/\,3.49$\times$ & 6.75\,/\,3.40$\times$ & 5.69\,/\,2.59$\times$ & 4.91\,/\,2.17$\times$ & 6.92\,/\,3.35$\times$ \\
\;\;$+$ Loop                             & 10.29\,/\,4.25$\times$ & 9.68\,/\,4.32$\times$ & 9.12\,/\,3.91$\times$ & 8.34\,/\,3.74$\times$ & 7.92\,/\,3.58$\times$ & 7.60\,/\,3.46$\times$ & 6.32\,/\,2.65$\times$ & 5.05\,/\,\textbf{2.27}$\times$ & 8.04\,/\,3.52$\times$ \\
\;\;$+$ Train                            & 8.11\,/\,3.99$\times$ & 8.02\,/\,4.06$\times$ & 7.50\,/\,3.60$\times$ & 7.31\,/\,3.59$\times$ & 7.32\,/\,3.49$\times$ & 6.97\,/\,3.46$\times$ & 5.75\,/\,2.60$\times$ & 4.97\,/\,2.17$\times$ & 6.99\,/\,3.37$\times$ \\
\;\;$+$ \textbf{DLoop}                   & \textbf{14.14}\,/\,\textbf{4.79}$\times$ & \textbf{12.25}\,/\,\textbf{4.87}$\times$ & \textbf{9.73}\,/\,\textbf{4.09}$\times$ & \textbf{9.85}\,/\,\textbf{3.96}$\times$ & \textbf{8.94}\,/\,\textbf{3.81}$\times$ & \textbf{9.51}\,/\,\textbf{3.70}$\times$ & \textbf{7.22}\,/\,\textbf{2.69}$\times$ & \textbf{5.36}\,/\,2.25$\times$ & \textbf{9.63}\,/\,\textbf{3.77}$\times$ \\
\midrule
DFlash                                   & 6.51\,/\,4.60$\times$ & 7.79\,/\,5.52$\times$ & 7.26\,/\,4.94$\times$ & 6.69\,/\,4.84$\times$ & 6.12\,/\,4.23$\times$ & 6.97\,/\,4.88$\times$ & 4.38\,/\,2.58$\times$ & 3.09\,/\,1.95$\times$ & 6.10\,/\,4.19$\times$ \\
\;\;$+$ Loop                             & 6.51\,/\,4.56$\times$ & 7.81\,/\,5.48$\times$ & 7.27\,/\,4.94$\times$ & 6.73\,/\,4.61$\times$ & 6.13\,/\,4.19$\times$ & 6.98\,/\,4.94$\times$ & 4.39\,/\,2.59$\times$ & 3.09\,/\,1.98$\times$ & 6.11\,/\,4.16$\times$ \\
\;\;$+$ Train                            & 8.56\,/\,5.15$\times$ & 8.31\,/\,5.63$\times$ & 7.12\,/\,4.79$\times$ & 6.37\,/\,4.59$\times$ & 5.98\,/\,4.11$\times$ & 6.91\,/\,4.82$\times$ & 4.62\,/\,2.72$\times$ & 3.20\,/\,2.05$\times$ & 6.38\,/\,4.23$\times$ \\
\;\;$+$ \textbf{DLoop}                   & 10.43\,/\,6.73$\times$ & 9.99\,/\,6.46$\times$ & 8.02\,/\,5.21$\times$ & 6.97\,/\,4.82$\times$ & 6.32\,/\,4.26$\times$ & 7.68\,/\,5.07$\times$ & 5.00\,/\,2.67$\times$ & 3.24\,/\,2.09$\times$ & 7.21\,/\,4.67$\times$ \\
\;\;$+$ DDTree                           & 9.05\,/\,6.33$\times$ & 10.36\,/\,7.45$\times$ & 9.58\,/\,6.42$\times$ & 9.37\,/\,6.57$\times$ & 8.85\,/\,6.03$\times$ & 9.47\,/\,6.40$\times$ & 6.51\,/\,3.94$\times$ & 4.82\,/\,3.18$\times$ & 8.50\,/\,5.79$\times$ \\
\;\;$+$ DDTree $+$ \textbf{DLoop}        & \textbf{14.06}\,/\,\textbf{8.59}$\times$ & \textbf{13.94}\,/\,\textbf{8.52}$\times$ & \textbf{11.81}\,/\,\textbf{7.23}$\times$ & \textbf{10.38}\,/\,\textbf{6.93}$\times$ & \textbf{9.48}\,/\,\textbf{6.07}$\times$ & \textbf{10.75}\,/\,\textbf{6.87}$\times$ & \textbf{7.37}\,/\,\textbf{4.19}$\times$ & \textbf{5.17}\,/\,\textbf{3.43}$\times$ & \textbf{10.37}\,/\,\textbf{6.48}$\times$ \\
\midrule
Domino                                   & 10.08\,/\,6.40$\times$ & 8.99\,/\,5.83$\times$ & 7.48\,/\,4.67$\times$ & 7.07\,/\,4.49$\times$ & 7.03\,/\,4.28$\times$ & 6.86\,/\,4.39$\times$ & 5.26\,/\,2.75$\times$ & 3.87\,/\,2.21$\times$ & 7.08\,/\,4.38$\times$ \\
\;\;$+$ Loop                             & 10.32\,/\,6.19$\times$ & 9.15\,/\,5.59$\times$ & 7.55\,/\,4.62$\times$ & 7.11\,/\,4.53$\times$ & 7.06\,/\,4.33$\times$ & 6.90\,/\,4.37$\times$ & 5.27\,/\,2.73$\times$ & 3.88\,/\,2.21$\times$ & 7.15\,/\,4.32$\times$ \\
\;\;$+$ Train                            & 9.99\,/\,6.35$\times$ & 9.05\,/\,5.79$\times$ & 7.45\,/\,4.65$\times$ & 7.00\,/\,4.60$\times$ & 6.95\,/\,4.26$\times$ & 6.98\,/\,4.45$\times$ & 5.25\,/\,2.70$\times$ & 3.84\,/\,2.20$\times$ & 7.06\,/\,4.38$\times$ \\
\;\;$+$ \textbf{DLoop}                   & 12.80\,/\,6.98$\times$ & 10.72\,/\,6.31$\times$ & 7.81\,/\,4.66$\times$ & 7.51\,/\,4.60$\times$ & 7.33\,/\,4.38$\times$ & 7.40\,/\,4.65$\times$ & 5.69\,/\,2.72$\times$ & 3.95\,/\,2.38$\times$ & 7.90\,/\,4.58$\times$ \\
\;\;$+$ DARTree                          & 12.98\,/\,6.71$\times$ & 12.04\,/\,6.48$\times$ & 10.31\,/\,5.27$\times$ & 9.99\,/\,5.38$\times$ & 10.21\,/\,5.30$\times$ & 9.31\,/\,5.09$\times$ & 7.38\,/\,3.42$\times$ & 5.78\,/\,\textbf{2.87}$\times$ & 9.75\,/\,5.06$\times$ \\
\;\;$+$ DARTree $+$ \textbf{DLoop}       & \textbf{16.65}\,/\,\textbf{7.65}$\times$ & \textbf{14.47}\,/\,\textbf{7.12}$\times$ & \textbf{11.55}\,/\,\textbf{5.73}$\times$ & \textbf{11.08}\,/\,\textbf{5.58}$\times$ & \textbf{10.88}\,/\,\textbf{5.59}$\times$ & \textbf{10.47}\,/\,\textbf{5.13}$\times$ & \textbf{8.14}\,/\,\textbf{3.50}$\times$ & \textbf{6.10}\,/\,2.86$\times$ & \textbf{11.17}\,/\,\textbf{5.40}$\times$ \\
\midrule
DSpark                                   & 6.58\,/\,4.55$\times$ & 6.44\,/\,4.62$\times$ & 5.75\,/\,3.90$\times$ & 5.73\,/\,4.05$\times$ & 5.57\,/\,3.86$\times$ & 5.55\,/\,3.97$\times$ & 4.48\,/\,2.82$\times$ & 3.70\,/\,2.42$\times$ & 5.47\,/\,3.77$\times$ \\
\;\;$+$ Loop                             & 7.20\,/\,4.62$\times$ & 6.91\,/\,4.73$\times$ & 6.01\,/\,3.99$\times$ & 6.03\,/\,4.08$\times$ & 5.82\,/\,3.85$\times$ & 5.80\,/\,3.82$\times$ & 4.63\,/\,2.80$\times$ & 3.74\,/\,2.40$\times$ & 5.77\,/\,3.79$\times$ \\
\;\;$+$ Train                            & 6.53\,/\,4.52$\times$ & 6.33\,/\,4.56$\times$ & 5.63\,/\,3.86$\times$ & 5.59\,/\,4.01$\times$ & 5.50\,/\,3.80$\times$ & 5.66\,/\,4.01$\times$ & 4.40\,/\,2.74$\times$ & 3.64\,/\,2.35$\times$ & 5.41\,/\,3.73$\times$ \\
\;\;$+$ \textbf{DLoop}                   & 13.34\,/\,6.69$\times$ & 11.38\,/\,6.39$\times$ & 8.34\,/\,\textbf{4.97}$\times$ & 8.31\,/\,5.00$\times$ & 7.81\,/\,4.50$\times$ & 9.19\,/\,5.08$\times$ & 6.27\,/\,2.95$\times$ & 4.08\,/\,2.46$\times$ & 8.59\,/\,4.76$\times$ \\
\;\;$+$ DARTree                          & 7.27\,/\,4.84$\times$ & 7.04\,/\,4.79$\times$ & 6.50\,/\,4.19$\times$ & 6.94\,/\,4.62$\times$ & 6.77\,/\,4.32$\times$ & 6.63\,/\,4.48$\times$ & 5.55\,/\,3.48$\times$ & 4.90\,/\,3.03$\times$ & 6.45\,/\,4.22$\times$ \\
\;\;$+$ DARTree $+$ \textbf{DLoop}       & \textbf{15.41}\,/\,\textbf{6.80}$\times$ & \textbf{13.02}\,/\,\textbf{6.56}$\times$ & \textbf{9.38}\,/\,4.93$\times$ & \textbf{10.45}\,/\,\textbf{5.71}$\times$ & \textbf{9.38}\,/\,\textbf{5.01}$\times$ & \textbf{10.40}\,/\,\textbf{5.50}$\times$ & \textbf{7.66}\,/\,\textbf{3.65}$\times$ & \textbf{5.38}\,/\,\textbf{3.04}$\times$ & \textbf{10.14}\,/\,\textbf{5.15}$\times$ \\
\midrule
\multicolumn{10}{c}{\textbf{Qwen3-8B}} \\
\midrule
EAGLE-3                                  & 7.87\,/\,3.94$\times$ & 7.81\,/\,3.82$\times$ & 7.10\,/\,3.58$\times$ & 6.79\,/\,3.47$\times$ & 6.85\,/\,3.29$\times$ & 6.55\,/\,3.14$\times$ & 5.24\,/\,2.40$\times$ & 4.72\,/\,2.13$\times$ & 6.61\,/\,3.22$\times$ \\
\;\;$+$ Loop                             & 9.43\,/\,4.30$\times$ & 9.33\,/\,4.10$\times$ & 8.16\,/\,3.89$\times$ & 7.32\,/\,3.53$\times$ & 7.12\,/\,3.37$\times$ & 7.05\,/\,3.21$\times$ & 5.48\,/\,\textbf{2.42}$\times$ & 4.80\,/\,\textbf{2.23}$\times$ & 7.34\,/\,3.38$\times$ \\
\;\;$+$ Train                            & 7.90\,/\,3.84$\times$ & 7.80\,/\,3.82$\times$ & 6.97\,/\,3.55$\times$ & 6.80\,/\,3.39$\times$ & 6.86\,/\,3.23$\times$ & 6.53\,/\,3.12$\times$ & 5.26\,/\,2.37$\times$ & 4.71\,/\,1.87$\times$ & 6.60\,/\,3.15$\times$ \\
\;\;$+$ \textbf{DLoop}                   & \textbf{11.76}\,/\,\textbf{4.62}$\times$ & \textbf{10.98}\,/\,\textbf{4.29}$\times$ & \textbf{8.61}\,/\,\textbf{4.01}$\times$ & \textbf{7.95}\,/\,\textbf{3.66}$\times$ & \textbf{7.77}\,/\,\textbf{3.45}$\times$ & \textbf{7.77}\,/\,\textbf{3.28}$\times$ & \textbf{6.03}\,/\,2.51$\times$ & \textbf{4.90}\,/\,2.25$\times$ & \textbf{8.22}\,/\,\textbf{3.51}$\times$ \\
\midrule
DFlash                                   & 6.55\,/\,4.68$\times$ & 7.93\,/\,5.50$\times$ & 7.18\,/\,5.22$\times$ & 6.56\,/\,4.47$\times$ & 5.98\,/\,4.05$\times$ & 7.16\,/\,4.82$\times$ & 4.23\,/\,2.45$\times$ & 3.09\,/\,1.99$\times$ & 6.08\,/\,4.15$\times$ \\
\;\;$+$ Loop                             & 6.60\,/\,4.64$\times$ & 8.11\,/\,5.29$\times$ & 7.59\,/\,5.45$\times$ & 6.60\,/\,4.61$\times$ & 6.04\,/\,4.11$\times$ & 7.29\,/\,4.81$\times$ & 4.27\,/\,2.46$\times$ & 3.10\,/\,2.02$\times$ & 6.20\,/\,4.17$\times$ \\
\;\;$+$ Train                            & 8.64\,/\,5.48$\times$ & 8.69\,/\,5.86$\times$ & 7.09\,/\,5.08$\times$ & 6.50\,/\,4.60$\times$ & 6.06\,/\,4.19$\times$ & 7.17\,/\,4.76$\times$ & 4.51\,/\,2.59$\times$ & 3.19\,/\,2.09$\times$ & 6.48\,/\,4.33$\times$ \\
\;\;$+$ \textbf{DLoop}                   & 10.73\,/\,6.78$\times$ & 10.79\,/\,6.60$\times$ & 7.83\,/\,5.46$\times$ & 7.17\,/\,4.78$\times$ & 6.36\,/\,4.27$\times$ & 8.12\,/\,5.06$\times$ & 4.93\,/\,2.65$\times$ & 3.24\,/\,2.08$\times$ & 7.40\,/\,4.71$\times$ \\
\;\;$+$ DDTree                           & 9.12\,/\,6.55$\times$ & 10.46\,/\,7.27$\times$ & 9.39\,/\,6.65$\times$ & 9.33\,/\,6.73$\times$ & 8.66\,/\,5.93$\times$ & 9.75\,/\,6.66$\times$ & 6.30\,/\,3.81$\times$ & 4.82\,/\,3.34$\times$ & 8.48\,/\,5.87$\times$ \\
\;\;$+$ DDTree $+$ \textbf{DLoop}        & \textbf{14.24}\,/\,\textbf{9.08}$\times$ & \textbf{14.14}\,/\,\textbf{8.68}$\times$ & \textbf{10.54}\,/\,\textbf{7.31}$\times$ & \textbf{10.38}\,/\,\textbf{6.97}$\times$ & \textbf{9.10}\,/\,\textbf{6.26}$\times$ & \textbf{11.07}\,/\,\textbf{7.11}$\times$ & \textbf{7.18}\,/\,\textbf{4.06}$\times$ & \textbf{5.07}\,/\,\textbf{3.52}$\times$ & \textbf{10.22}\,/\,\textbf{6.62}$\times$ \\
\midrule
Domino                                   & 9.98\,/\,6.39$\times$ & 9.43\,/\,5.80$\times$ & 7.72\,/\,5.05$\times$ & 7.33\,/\,4.75$\times$ & 7.11\,/\,4.37$\times$ & 7.08\,/\,4.35$\times$ & 5.19\,/\,2.70$\times$ & 3.90\,/\,2.32$\times$ & 7.22\,/\,4.47$\times$ \\
\;\;$+$ Loop                             & 10.56\,/\,6.47$\times$ & 9.87\,/\,5.87$\times$ & 7.76\,/\,5.04$\times$ & 7.44\,/\,4.68$\times$ & 7.17\,/\,4.44$\times$ & 7.20\,/\,4.45$\times$ & 5.26\,/\,2.76$\times$ & 3.93\,/\,2.35$\times$ & 7.40\,/\,4.51$\times$ \\
\;\;$+$ Train                            & 10.03\,/\,6.57$\times$ & 9.47\,/\,5.80$\times$ & 7.73\,/\,5.13$\times$ & 7.36\,/\,4.86$\times$ & 7.06\,/\,4.43$\times$ & 7.10\,/\,4.33$\times$ & 5.19\,/\,2.70$\times$ & 3.90\,/\,2.35$\times$ & 7.23\,/\,4.52$\times$ \\
\;\;$+$ \textbf{DLoop}                   & 13.42\,/\,7.30$\times$ & 11.70\,/\,6.12$\times$ & 8.32\,/\,5.52$\times$ & 7.97\,/\,4.93$\times$ & 7.33\,/\,4.46$\times$ & 7.78\,/\,4.54$\times$ & 5.76\,/\,2.77$\times$ & 4.06\,/\,2.45$\times$ & 8.29\,/\,4.76$\times$ \\
\;\;$+$ DARTree                          & 12.94\,/\,6.97$\times$ & 12.19\,/\,6.32$\times$ & 10.45\,/\,5.65$\times$ & 10.35\,/\,5.61$\times$ & 10.10\,/\,5.26$\times$ & 9.73\,/\,5.00$\times$ & 7.38\,/\,3.31$\times$ & 5.80\,/\,2.89$\times$ & 9.87\,/\,5.13$\times$ \\
\;\;$+$ DARTree $+$ \textbf{DLoop}       & \textbf{17.23}\,/\,\textbf{7.80}$\times$ & \textbf{15.44}\,/\,\textbf{7.00}$\times$ & \textbf{12.57}\,/\,\textbf{6.19}$\times$ & \textbf{11.93}\,/\,\textbf{5.87}$\times$ & \textbf{11.06}\,/\,\textbf{5.63}$\times$ & \textbf{11.16}\,/\,\textbf{5.26}$\times$ & \textbf{8.25}\,/\,\textbf{3.43}$\times$ & \textbf{6.17}\,/\,\textbf{2.97}$\times$ & \textbf{11.73}\,/\,\textbf{5.52}$\times$ \\
\midrule
DSpark                                   & 6.64\,/\,4.74$\times$ & 6.55\,/\,4.51$\times$ & 5.79\,/\,4.14$\times$ & 5.88\,/\,4.08$\times$ & 5.72\,/\,3.81$\times$ & 5.83\,/\,3.98$\times$ & 4.41\,/\,2.72$\times$ & 3.85\,/\,2.49$\times$ & 5.58\,/\,3.81$\times$ \\
\;\;$+$ Loop                             & 7.27\,/\,4.86$\times$ & 7.10\,/\,4.47$\times$ & 6.20\,/\,4.28$\times$ & 6.16\,/\,4.16$\times$ & 5.95\,/\,3.87$\times$ & 6.24\,/\,3.95$\times$ & 4.54\,/\,2.78$\times$ & 3.90\,/\,2.55$\times$ & 5.92\,/\,3.87$\times$ \\
\;\;$+$ Train                            & 6.60\,/\,4.67$\times$ & 6.42\,/\,4.42$\times$ & 5.70\,/\,4.05$\times$ & 5.76\,/\,4.05$\times$ & 5.60\,/\,3.82$\times$ & 5.81\,/\,4.04$\times$ & 4.37\,/\,2.63$\times$ & 3.77\,/\,2.49$\times$ & 5.50\,/\,3.77$\times$ \\
\;\;$+$ \textbf{DLoop}                   & 13.83\,/\,6.94$\times$ & 12.66\,/\,6.08$\times$ & 8.90\,/\,\textbf{5.58}$\times$ & 9.03\,/\,5.22$\times$ & 7.99\,/\,4.58$\times$ & 9.48\,/\,4.70$\times$ & 6.29\,/\,2.91$\times$ & 4.49\,/\,2.57$\times$ & 9.08\,/\,4.82$\times$ \\
\;\;$+$ DARTree                          & 7.43\,/\,4.86$\times$ & 7.21\,/\,4.50$\times$ & 6.61\,/\,4.37$\times$ & 7.04\,/\,4.58$\times$ & 6.89\,/\,4.35$\times$ & 6.85\,/\,4.28$\times$ & 5.60\,/\,3.39$\times$ & 5.24\,/\,\textbf{3.31}$\times$ & 6.61\,/\,4.21$\times$ \\
\;\;$+$ DARTree $+$ \textbf{DLoop}       & \textbf{14.20}\,/\,\textbf{7.25}$\times$ & \textbf{13.00}\,/\,\textbf{6.31}$\times$ & \textbf{9.73}\,/\,5.51$\times$ & \textbf{10.60}\,/\,\textbf{5.85}$\times$ & \textbf{9.23}\,/\,\textbf{5.00}$\times$ & \textbf{10.65}\,/\,\textbf{5.42}$\times$ & \textbf{7.71}\,/\,\textbf{3.60}$\times$ & \textbf{5.74}\,/\,\textbf{3.31}$\times$ & \textbf{10.11}\,/\,\textbf{5.28}$\times$ \\
\midrule
\multicolumn{10}{c}{\textbf{Qwen3.5-9B}} \\
\midrule
MTP                                      & 3.59\,/\,1.76$\times$ & 3.59\,/\,1.80$\times$ & 3.45\,/\,1.64$\times$ & 3.60\,/\,1.71$\times$ & 3.30\,/\,1.53$\times$ & 3.27\,/\,1.49$\times$ & 2.98\,/\,1.30$\times$ & 2.78\,/\,1.13$\times$ & 3.32\,/\,1.55$\times$ \\
\;\;$+$ Loop                             & 6.04\,/\,2.19$\times$ & 5.82\,/\,2.20$\times$ & 5.12\,/\,1.91$\times$ & 6.73\,/\,2.25$\times$ & 4.35\,/\,1.69$\times$ & 4.43\,/\,1.63$\times$ & 4.02\,/\,1.38$\times$ & 3.12\,/\,1.16$\times$ & 4.96\,/\,1.80$\times$ \\
\;\;$+$ Train                            & 3.72\,/\,1.87$\times$ & 3.66\,/\,1.85$\times$ & 3.45\,/\,1.63$\times$ & 3.66\,/\,1.80$\times$ & 3.42\,/\,1.65$\times$ & 3.29\,/\,1.53$\times$ & 2.99\,/\,1.30$\times$ & 2.81\,/\,1.16$\times$ & 3.37\,/\,1.60$\times$ \\
\;\;$+$ \textbf{DLoop}                   & \textbf{10.46}\,/\,\textbf{3.13}$\times$ & \textbf{8.39}\,/\,\textbf{2.75}$\times$ & \textbf{5.92}\,/\,\textbf{2.09}$\times$ & \textbf{9.47}\,/\,\textbf{2.83}$\times$ & \textbf{5.72}\,/\,\textbf{2.09}$\times$ & \textbf{5.59}\,/\,\textbf{1.87}$\times$ & \textbf{4.63}\,/\,\textbf{1.44}$\times$ & \textbf{3.41}\,/\,\textbf{1.22}$\times$ & \textbf{6.70}\,/\,\textbf{2.18}$\times$ \\
\midrule
\multicolumn{10}{c}{\textbf{Gemma-4-12B}} \\
\midrule
MTP                                      & 5.45\,/\,3.57$\times$ & 5.48\,/\,3.71$\times$ & 5.20\,/\,3.38$\times$ & 5.63\,/\,3.75$\times$ & 4.72\,/\,3.13$\times$ & 5.01\,/\,3.24$\times$ & 3.91\,/\,2.41$\times$ & 3.51\,/\,\textbf{2.16}$\times$ & 4.86\,/\,3.17$\times$ \\
\;\;$+$ Loop                             & 7.14\,/\,3.28$\times$ & 7.11\,/\,3.30$\times$ & 6.64\,/\,3.02$\times$ & 7.88\,/\,3.38$\times$ & 5.49\,/\,2.96$\times$ & 6.06\,/\,2.88$\times$ & 4.52\,/\,2.31$\times$ & 3.68\,/\,2.10$\times$ & 6.06\,/\,2.90$\times$ \\
\;\;$+$ Train                            & 5.83\,/\,3.81$\times$ & 5.65\,/\,3.77$\times$ & 5.13\,/\,3.34$\times$ & 5.74\,/\,3.76$\times$ & 4.90\,/\,3.40$\times$ & 4.99\,/\,3.23$\times$ & 3.88\,/\,2.36$\times$ & 3.45\,/\,2.07$\times$ & 4.95\,/\,3.22$\times$ \\
\;\;$+$ \textbf{DLoop}                   & \textbf{10.96}\,/\,\textbf{4.80}$\times$ & \textbf{9.23}\,/\,\textbf{4.48}$\times$ & \textbf{6.99}\,/\,\textbf{3.69}$\times$ & \textbf{10.61}\,/\,\textbf{4.47}$\times$ & \textbf{6.42}\,/\,\textbf{3.80}$\times$ & \textbf{6.99}\,/\,\textbf{3.57}$\times$ & \textbf{4.97}\,/\,\textbf{2.45}$\times$ & \textbf{3.82}\,/\,2.15$\times$ & \textbf{7.50}\,/\,\textbf{3.68}$\times$ \\
\bottomrule
\end{tabular}}
\end{table}

\begin{table}[p]
\centering
\caption{\textbf{Acceptance length ($\tau$) and speedup (spd) on each benchmark with sampling at temperature $T=1$.}
The rows and the notation follow Table~\ref{tab:full}. Verification follows the sampling rule of the official implementation of each draft model. The gate threshold is selected at $T=1$ in the same way as at $T=0$. Bold marks the best result in each column for each draft model.}
\label{tab:full_t1}
\setlength{\tabcolsep}{2.5pt}
\renewcommand{\arraystretch}{1.0}
\resizebox{\textwidth}{!}{\footnotesize
\begin{tabular}{@{}l*{9}{r}@{}}
\toprule
 & \multicolumn{3}{c}{\textsc{Math}} & \multicolumn{3}{c}{\textsc{Code}} & \multicolumn{2}{c}{\textsc{Chat}} & \\
\cmidrule(lr){2-4} \cmidrule(lr){5-7} \cmidrule(lr){8-9}
Method & \multicolumn{1}{c}{GSM8K} & \multicolumn{1}{c}{MATH-500} & \multicolumn{1}{c}{AIME25} & \multicolumn{1}{c}{HumanEval} & \multicolumn{1}{c}{MBPP} & \multicolumn{1}{c}{LCB} & \multicolumn{1}{c}{MT-Bench} & \multicolumn{1}{c}{Alpaca} & \multicolumn{1}{c}{\textit{Avg.}} \\
\midrule
\multicolumn{10}{c}{\textbf{Qwen3-4B}} \\
\midrule
EAGLE-3                                  & 7.83\,/\,3.80$\times$ & 7.36\,/\,3.59$\times$ & 5.97\,/\,2.98$\times$ & 6.94\,/\,3.39$\times$ & 6.99\,/\,3.37$\times$ & 6.56\,/\,3.16$\times$ & 5.27\,/\,2.32$\times$ & 4.69\,/\,1.80$\times$ & 6.45\,/\,3.05$\times$ \\
\;\;$+$ Loop                             & 9.71\,/\,4.02$\times$ & 8.68\,/\,3.76$\times$ & 6.54\,/\,3.13$\times$ & 8.01\,/\,3.55$\times$ & 7.53\,/\,3.49$\times$ & 6.95\,/\,3.12$\times$ & 5.68\,/\,2.32$\times$ & 4.85\,/\,2.07$\times$ & 7.24\,/\,3.18$\times$ \\
\;\;$+$ Train                            & 7.91\,/\,3.86$\times$ & 7.32\,/\,3.55$\times$ & 6.18\,/\,3.09$\times$ & 6.95\,/\,3.42$\times$ & 7.07\,/\,3.50$\times$ & 6.75\,/\,3.24$\times$ & 5.43\,/\,2.43$\times$ & 4.79\,/\,2.07$\times$ & 6.55\,/\,3.14$\times$ \\
\;\;$+$ \textbf{DLoop}                   & \textbf{12.86}\,/\,\textbf{4.46}$\times$ & \textbf{10.95}\,/\,\textbf{4.09}$\times$ & \textbf{7.08}\,/\,\textbf{3.23}$\times$ & \textbf{9.10}\,/\,\textbf{3.76}$\times$ & \textbf{8.63}\,/\,\textbf{3.77}$\times$ & \textbf{8.30}\,/\,\textbf{3.42}$\times$ & \textbf{6.55}\,/\,\textbf{2.48}$\times$ & \textbf{5.05}\,/\,\textbf{2.10}$\times$ & \textbf{8.56}\,/\,\textbf{3.41}$\times$ \\
\midrule
DFlash                                   & 6.03\,/\,4.32$\times$ & 6.59\,/\,4.61$\times$ & 4.90\,/\,3.29$\times$ & 6.00\,/\,4.16$\times$ & 5.61\,/\,3.99$\times$ & 6.40\,/\,4.35$\times$ & 4.01\,/\,2.37$\times$ & 2.91\,/\,1.81$\times$ & 5.31\,/\,3.61$\times$ \\
\;\;$+$ Loop                             & 6.03\,/\,4.23$\times$ & 6.83\,/\,4.50$\times$ & 5.11\,/\,3.59$\times$ & 6.06\,/\,4.19$\times$ & 5.68\,/\,3.84$\times$ & 6.66\,/\,4.46$\times$ & 4.07\,/\,2.31$\times$ & 2.92\,/\,1.84$\times$ & 5.42\,/\,3.62$\times$ \\
\;\;$+$ Train                            & 7.46\,/\,5.32$\times$ & 6.91\,/\,4.75$\times$ & 4.72\,/\,3.32$\times$ & 5.86\,/\,4.19$\times$ & 5.58\,/\,3.96$\times$ & 6.59\,/\,4.46$\times$ & 4.18\,/\,2.42$\times$ & 2.99\,/\,1.95$\times$ & 5.54\,/\,3.80$\times$ \\
\;\;$+$ \textbf{DLoop}                   & 8.75\,/\,5.53$\times$ & 7.87\,/\,5.07$\times$ & 5.03\,/\,3.42$\times$ & 6.24\,/\,4.39$\times$ & 5.67\,/\,3.95$\times$ & 6.69\,/\,4.35$\times$ & 4.46\,/\,2.39$\times$ & 3.03\,/\,1.99$\times$ & 5.97\,/\,3.89$\times$ \\
\;\;$+$ DDTree                           & 8.63\,/\,6.01$\times$ & 9.27\,/\,6.16$\times$ & 7.45\,/\,\textbf{4.99}$\times$ & 8.71\,/\,\textbf{6.02}$\times$ & 8.39\,/\,\textbf{5.76}$\times$ & 9.00\,/\,5.79$\times$ & 6.04\,/\,3.51$\times$ & 4.66\,/\,2.88$\times$ & 7.77\,/\,5.14$\times$ \\
\;\;$+$ DDTree $+$ \textbf{DLoop}        & \textbf{11.58}\,/\,\textbf{7.46}$\times$ & \textbf{10.84}\,/\,\textbf{6.75}$\times$ & \textbf{7.74}\,/\,4.90$\times$ & \textbf{9.29}\,/\,5.99$\times$ & \textbf{8.52}\,/\,5.73$\times$ & \textbf{9.43}\,/\,\textbf{6.12}$\times$ & \textbf{6.61}\,/\,\textbf{3.66}$\times$ & \textbf{4.92}\,/\,\textbf{3.12}$\times$ & \textbf{8.62}\,/\,\textbf{5.47}$\times$ \\
\midrule
Domino                                   & 7.98\,/\,4.69$\times$ & 6.87\,/\,3.88$\times$ & 4.19\,/\,2.51$\times$ & 5.66\,/\,3.36$\times$ & 5.50\,/\,3.20$\times$ & 5.79\,/\,3.28$\times$ & 4.11\,/\,1.98$\times$ & 3.11\,/\,1.64$\times$ & 5.40\,/\,3.07$\times$ \\
\;\;$+$ Loop                             & 8.36\,/\,4.65$\times$ & 7.08\,/\,3.83$\times$ & 4.35\,/\,2.56$\times$ & 5.75\,/\,3.34$\times$ & 5.53\,/\,3.20$\times$ & 5.83\,/\,3.23$\times$ & 4.10\,/\,1.97$\times$ & 3.13\,/\,1.64$\times$ & 5.51\,/\,3.05$\times$ \\
\;\;$+$ Train                            & 7.96\,/\,4.65$\times$ & 6.87\,/\,3.92$\times$ & 4.25\,/\,2.52$\times$ & 5.65\,/\,3.36$\times$ & 5.47\,/\,3.17$\times$ & 5.60\,/\,3.27$\times$ & 3.97\,/\,1.91$\times$ & 3.07\,/\,1.64$\times$ & 5.36\,/\,3.05$\times$ \\
\;\;$+$ \textbf{DLoop}                   & 9.58\,/\,4.98$\times$ & 7.76\,/\,3.95$\times$ & 4.45\,/\,2.59$\times$ & 5.92\,/\,3.39$\times$ & 5.60\,/\,3.17$\times$ & 5.86\,/\,3.37$\times$ & 4.16\,/\,1.89$\times$ & 3.14\,/\,1.62$\times$ & 5.81\,/\,3.12$\times$ \\
\;\;$+$ DARTree                          & 11.50\,/\,5.96$\times$ & 9.93\,/\,5.13$\times$ & 7.28\,/\,3.84$\times$ & 9.14\,/\,4.86$\times$ & 9.25\,/\,4.83$\times$ & 8.93\,/\,4.57$\times$ & 6.75\,/\,3.06$\times$ & 5.41\,/\,2.55$\times$ & 8.52\,/\,4.35$\times$ \\
\;\;$+$ DARTree $+$ \textbf{DLoop}       & \textbf{14.12}\,/\,\textbf{6.55}$\times$ & \textbf{11.62}\,/\,\textbf{5.36}$\times$ & \textbf{8.10}\,/\,\textbf{4.04}$\times$ & \textbf{10.15}\,/\,\textbf{5.03}$\times$ & \textbf{10.07}\,/\,\textbf{5.10}$\times$ & \textbf{9.45}\,/\,\textbf{4.69}$\times$ & \textbf{7.34}\,/\,\textbf{3.14}$\times$ & \textbf{5.69}\,/\,\textbf{2.57}$\times$ & \textbf{9.57}\,/\,\textbf{4.56}$\times$ \\
\midrule
DSpark                                   & 6.39\,/\,4.44$\times$ & 6.07\,/\,4.20$\times$ & 5.16\,/\,3.67$\times$ & 5.56\,/\,3.84$\times$ & 5.47\,/\,3.63$\times$ & 5.34\,/\,3.72$\times$ & 4.25\,/\,2.63$\times$ & 3.52\,/\,2.28$\times$ & 5.22\,/\,3.55$\times$ \\
\;\;$+$ Loop                             & 7.02\,/\,4.44$\times$ & 6.52\,/\,4.15$\times$ & 5.14\,/\,3.48$\times$ & 5.78\,/\,3.73$\times$ & 5.68\,/\,3.55$\times$ & 5.56\,/\,3.59$\times$ & 4.45\,/\,2.63$\times$ & 3.63\,/\,2.28$\times$ & 5.47\,/\,3.48$\times$ \\
\;\;$+$ Train                            & 6.32\,/\,4.45$\times$ & 5.93\,/\,4.14$\times$ & 4.94\,/\,3.55$\times$ & 5.41\,/\,3.71$\times$ & 5.33\,/\,3.67$\times$ & 5.39\,/\,3.64$\times$ & 4.21\,/\,2.59$\times$ & 3.53\,/\,2.27$\times$ & 5.13\,/\,3.50$\times$ \\
\;\;$+$ \textbf{DLoop}                   & \textbf{11.71}\,/\,6.07$\times$ & \textbf{10.40}\,/\,\textbf{5.37}$\times$ & 6.45\,/\,3.97$\times$ & 7.61\,/\,4.41$\times$ & 7.03\,/\,4.12$\times$ & 7.52\,/\,4.09$\times$ & 5.51\,/\,2.72$\times$ & 3.95\,/\,2.30$\times$ & 7.52\,/\,4.13$\times$ \\
\;\;$+$ DARTree                          & 7.04\,/\,4.71$\times$ & 6.55\,/\,4.28$\times$ & 5.45\,/\,3.55$\times$ & 6.68\,/\,4.38$\times$ & 6.55\,/\,4.18$\times$ & 6.46\,/\,4.22$\times$ & 5.22\,/\,3.09$\times$ & 4.62\,/\,2.73$\times$ & 6.07\,/\,3.89$\times$ \\
\;\;$+$ DARTree $+$ \textbf{DLoop}       & 11.31\,/\,\textbf{6.28}$\times$ & 9.81\,/\,5.26$\times$ & \textbf{6.53}\,/\,\textbf{4.05}$\times$ & \textbf{9.11}\,/\,\textbf{5.26}$\times$ & \textbf{8.08}\,/\,\textbf{4.83}$\times$ & \textbf{8.94}\,/\,\textbf{4.87}$\times$ & \textbf{6.64}\,/\,\textbf{3.26}$\times$ & \textbf{5.05}\,/\,\textbf{2.90}$\times$ & \textbf{8.18}\,/\,\textbf{4.59}$\times$ \\
\midrule
\multicolumn{10}{c}{\textbf{Qwen3-8B}} \\
\midrule
EAGLE-3                                  & 7.54\,/\,3.58$\times$ & 7.15\,/\,3.41$\times$ & 5.72\,/\,2.78$\times$ & 6.22\,/\,3.00$\times$ & 6.40\,/\,3.10$\times$ & 6.37\,/\,2.96$\times$ & 4.91\,/\,2.17$\times$ & 4.45\,/\,1.83$\times$ & 6.10\,/\,2.85$\times$ \\
\;\;$+$ Loop                             & 8.87\,/\,3.82$\times$ & 8.03\,/\,3.56$\times$ & 6.12\,/\,2.92$\times$ & 6.63\,/\,3.07$\times$ & 6.58\,/\,3.12$\times$ & 6.69\,/\,3.01$\times$ & 5.10\,/\,2.17$\times$ & 4.52\,/\,1.97$\times$ & 6.57\,/\,2.96$\times$ \\
\;\;$+$ Train                            & 7.53\,/\,3.61$\times$ & 7.14\,/\,3.54$\times$ & 5.61\,/\,2.80$\times$ & 6.28\,/\,3.03$\times$ & 6.45\,/\,3.15$\times$ & 6.37\,/\,3.00$\times$ & 4.89\,/\,2.17$\times$ & 4.40\,/\,1.95$\times$ & 6.08\,/\,2.91$\times$ \\
\;\;$+$ \textbf{DLoop}                   & \textbf{10.37}\,/\,\textbf{4.09}$\times$ & \textbf{9.22}\,/\,\textbf{3.87}$\times$ & \textbf{6.35}\,/\,\textbf{2.95}$\times$ & \textbf{7.06}\,/\,\textbf{3.19}$\times$ & \textbf{7.06}\,/\,\textbf{3.27}$\times$ & \textbf{6.99}\,/\,\textbf{3.04}$\times$ & \textbf{5.43}\,/\,\textbf{2.22}$\times$ & \textbf{4.54}\,/\,\textbf{1.99}$\times$ & \textbf{7.13}\,/\,\textbf{3.08}$\times$ \\
\midrule
DFlash                                   & 5.95\,/\,4.09$\times$ & 6.47\,/\,4.50$\times$ & 4.84\,/\,3.28$\times$ & 5.53\,/\,3.92$\times$ & 5.23\,/\,3.69$\times$ & 6.79\,/\,4.38$\times$ & 3.85\,/\,2.26$\times$ & 2.99\,/\,1.88$\times$ & 5.21\,/\,3.50$\times$ \\
\;\;$+$ Loop                             & 5.95\,/\,4.09$\times$ & 6.47\,/\,4.27$\times$ & 4.98\,/\,3.23$\times$ & 5.58\,/\,3.89$\times$ & 5.27\,/\,3.63$\times$ & 6.86\,/\,4.28$\times$ & 3.77\,/\,2.16$\times$ & 2.97\,/\,1.88$\times$ & 5.23\,/\,3.43$\times$ \\
\;\;$+$ Train                            & 7.36\,/\,4.96$\times$ & 6.68\,/\,4.54$\times$ & 4.80\,/\,3.35$\times$ & 5.58\,/\,3.93$\times$ & 5.28\,/\,3.71$\times$ & 6.67\,/\,4.39$\times$ & 4.00\,/\,2.36$\times$ & 3.02\,/\,1.98$\times$ & 5.42\,/\,3.65$\times$ \\
\;\;$+$ \textbf{DLoop}                   & 8.88\,/\,5.38$\times$ & 8.22\,/\,4.85$\times$ & 4.93\,/\,3.22$\times$ & 5.99\,/\,4.08$\times$ & 5.45\,/\,3.59$\times$ & 7.41\,/\,4.46$\times$ & 4.41\,/\,2.36$\times$ & 3.07\,/\,1.91$\times$ & 6.05\,/\,3.73$\times$ \\
\;\;$+$ DDTree                           & 8.60\,/\,5.81$\times$ & 9.12\,/\,6.22$\times$ & 7.20\,/\,\textbf{4.74}$\times$ & 8.23\,/\,\textbf{5.65}$\times$ & 7.93\,/\,5.43$\times$ & 9.27\,/\,6.10$\times$ & 5.78\,/\,\textbf{3.53}$\times$ & 4.62\,/\,2.90$\times$ & 7.59\,/\,5.05$\times$ \\
\;\;$+$ DDTree $+$ \textbf{DLoop}        & \textbf{11.49}\,/\,\textbf{7.16}$\times$ & \textbf{10.22}\,/\,\textbf{6.44}$\times$ & \textbf{7.26}\,/\,4.62$\times$ & \textbf{8.61}\,/\,5.63$\times$ & \textbf{7.96}\,/\,\textbf{5.52}$\times$ & \textbf{9.88}\,/\,\textbf{6.29}$\times$ & \textbf{6.13}\,/\,3.52$\times$ & \textbf{4.78}\,/\,\textbf{3.12}$\times$ & \textbf{8.29}\,/\,\textbf{5.29}$\times$ \\
\midrule
Domino                                   & 8.08\,/\,4.74$\times$ & 6.76\,/\,3.89$\times$ & 4.33\,/\,2.59$\times$ & 5.43\,/\,3.29$\times$ & 5.18\,/\,3.01$\times$ & 6.06\,/\,3.37$\times$ & 3.90\,/\,1.91$\times$ & 3.04\,/\,1.66$\times$ & 5.35\,/\,3.06$\times$ \\
\;\;$+$ Loop                             & 8.26\,/\,4.53$\times$ & 6.88\,/\,3.85$\times$ & 4.46\,/\,2.60$\times$ & 5.38\,/\,3.20$\times$ & 5.22\,/\,3.00$\times$ & 6.10\,/\,3.33$\times$ & 3.97\,/\,1.90$\times$ & 3.00\,/\,1.66$\times$ & 5.41\,/\,3.01$\times$ \\
\;\;$+$ Train                            & 7.94\,/\,4.45$\times$ & 6.76\,/\,3.91$\times$ & 4.46\,/\,2.66$\times$ & 5.41\,/\,3.28$\times$ & 5.21\,/\,3.07$\times$ & 6.14\,/\,3.39$\times$ & 3.92\,/\,1.87$\times$ & 2.97\,/\,1.67$\times$ & 5.35\,/\,3.04$\times$ \\
\;\;$+$ \textbf{DLoop}                   & 9.09\,/\,4.57$\times$ & 7.57\,/\,3.89$\times$ & 4.48\,/\,2.62$\times$ & 5.54\,/\,3.25$\times$ & 5.34\,/\,3.09$\times$ & 6.33\,/\,3.39$\times$ & 4.10\,/\,1.87$\times$ & 2.99\,/\,1.67$\times$ & 5.68\,/\,3.04$\times$ \\
\;\;$+$ DARTree                          & 11.16\,/\,5.68$\times$ & 9.59\,/\,4.91$\times$ & 7.00\,/\,3.62$\times$ & 8.55\,/\,4.44$\times$ & 8.91\,/\,4.56$\times$ & 9.24\,/\,4.65$\times$ & 6.33\,/\,2.91$\times$ & 5.23\,/\,2.52$\times$ & 8.25\,/\,4.16$\times$ \\
\;\;$+$ DARTree $+$ \textbf{DLoop}       & \textbf{13.94}\,/\,\textbf{6.16}$\times$ & \textbf{11.60}\,/\,\textbf{5.27}$\times$ & \textbf{7.42}\,/\,\textbf{3.68}$\times$ & \textbf{9.62}\,/\,\textbf{4.68}$\times$ & \textbf{9.48}\,/\,\textbf{4.69}$\times$ & \textbf{10.19}\,/\,\textbf{4.84}$\times$ & \textbf{7.06}\,/\,\textbf{2.93}$\times$ & \textbf{5.60}\,/\,\textbf{2.60}$\times$ & \textbf{9.36}\,/\,\textbf{4.36}$\times$ \\
\midrule
DSpark                                   & 6.44\,/\,4.39$\times$ & 6.10\,/\,4.25$\times$ & 5.07\,/\,3.43$\times$ & 5.62\,/\,3.89$\times$ & 5.44\,/\,3.67$\times$ & 5.70\,/\,3.86$\times$ & 4.18\,/\,2.65$\times$ & 3.67\,/\,2.33$\times$ & 5.28\,/\,3.56$\times$ \\
\;\;$+$ Loop                             & 7.10\,/\,4.42$\times$ & 6.54\,/\,4.18$\times$ & 5.16\,/\,3.34$\times$ & 5.82\,/\,3.76$\times$ & 5.76\,/\,3.66$\times$ & 6.02\,/\,3.71$\times$ & 4.33\,/\,2.59$\times$ & 3.75\,/\,2.29$\times$ & 5.56\,/\,3.49$\times$ \\
\;\;$+$ Train                            & 6.40\,/\,4.30$\times$ & 5.99\,/\,4.11$\times$ & 4.95\,/\,3.45$\times$ & 5.44\,/\,3.71$\times$ & 5.35\,/\,3.63$\times$ & 5.54\,/\,3.67$\times$ & 4.14\,/\,2.64$\times$ & 3.63\,/\,2.28$\times$ & 5.18\,/\,3.47$\times$ \\
\;\;$+$ \textbf{DLoop}                   & 11.53\,/\,5.91$\times$ & 10.04\,/\,5.26$\times$ & 6.30\,/\,3.82$\times$ & 7.70\,/\,4.43$\times$ & 7.02\,/\,4.10$\times$ & 8.59\,/\,4.29$\times$ & 5.63\,/\,2.72$\times$ & 4.04\,/\,2.33$\times$ & 7.61\,/\,4.11$\times$ \\
\;\;$+$ DARTree                          & 7.21\,/\,4.57$\times$ & 6.72\,/\,4.31$\times$ & 5.61\,/\,3.54$\times$ & 6.50\,/\,4.00$\times$ & 6.63\,/\,4.16$\times$ & 6.77\,/\,4.12$\times$ & 5.17\,/\,3.11$\times$ & 4.97\,/\,\textbf{2.97}$\times$ & 6.20\,/\,3.85$\times$ \\
\;\;$+$ DARTree $+$ \textbf{DLoop}       & \textbf{12.09}\,/\,\textbf{6.04}$\times$ & \textbf{10.64}\,/\,\textbf{5.39}$\times$ & \textbf{6.71}\,/\,\textbf{3.85}$\times$ & \textbf{8.98}\,/\,\textbf{4.82}$\times$ & \textbf{8.19}\,/\,\textbf{4.58}$\times$ & \textbf{10.00}\,/\,\textbf{5.03}$\times$ & \textbf{6.53}\,/\,\textbf{3.21}$\times$ & \textbf{5.32}\,/\,2.89$\times$ & \textbf{8.56}\,/\,\textbf{4.48}$\times$ \\
\midrule
\multicolumn{10}{c}{\textbf{Qwen3.5-9B}} \\
\midrule
MTP                                      & 3.50\,/\,1.73$\times$ & 3.45\,/\,1.66$\times$ & 3.23\,/\,1.52$\times$ & 3.43\,/\,1.66$\times$ & 3.09\,/\,1.38$\times$ & 3.05\,/\,1.38$\times$ & 2.69\,/\,1.09$\times$ & 2.53\,/\,1.03$\times$ & 3.12\,/\,1.43$\times$ \\
\;\;$+$ Loop                             & 5.64\,/\,2.08$\times$ & 5.40\,/\,1.98$\times$ & 4.59\,/\,1.77$\times$ & 6.06\,/\,2.07$\times$ & 3.95\,/\,1.51$\times$ & 3.91\,/\,1.46$\times$ & 3.39\,/\,1.13$\times$ & 2.84\,/\,1.06$\times$ & 4.47\,/\,1.63$\times$ \\
\;\;$+$ Train                            & 3.57\,/\,1.76$\times$ & 3.51\,/\,1.73$\times$ & 3.26\,/\,1.53$\times$ & 3.51\,/\,1.73$\times$ & 3.16\,/\,1.45$\times$ & 3.08\,/\,1.39$\times$ & 2.69\,/\,1.08$\times$ & 2.54\,/\,1.05$\times$ & 3.17\,/\,1.47$\times$ \\
\;\;$+$ \textbf{DLoop}                   & \textbf{9.56}\,/\,\textbf{2.83}$\times$ & \textbf{7.43}\,/\,\textbf{2.40}$\times$ & \textbf{5.13}\,/\,\textbf{1.85}$\times$ & \textbf{7.87}\,/\,\textbf{2.39}$\times$ & \textbf{5.08}\,/\,\textbf{1.78}$\times$ & \textbf{4.88}\,/\,\textbf{1.62}$\times$ & \textbf{4.01}\,/\,\textbf{1.20}$\times$ & \textbf{3.08}\,/\,\textbf{1.10}$\times$ & \textbf{5.88}\,/\,\textbf{1.90}$\times$ \\
\midrule
\multicolumn{10}{c}{\textbf{Gemma-4-12B}} \\
\midrule
MTP                                      & 5.28\,/\,3.49$\times$ & 5.21\,/\,3.56$\times$ & 4.86\,/\,3.19$\times$ & 5.35\,/\,3.61$\times$ & 4.50\,/\,2.96$\times$ & 4.61\,/\,2.94$\times$ & 3.82\,/\,2.31$\times$ & 3.44\,/\,2.07$\times$ & 4.63\,/\,3.02$\times$ \\
\;\;$+$ Loop                             & 6.75\,/\,3.18$\times$ & 6.62\,/\,3.17$\times$ & 5.83\,/\,2.80$\times$ & 7.35\,/\,3.27$\times$ & 5.17\,/\,2.79$\times$ & 5.49\,/\,2.60$\times$ & 4.40\,/\,2.18$\times$ & 3.57\,/\,2.02$\times$ & 5.65\,/\,2.75$\times$ \\
\;\;$+$ Train                            & 5.59\,/\,3.76$\times$ & 5.33\,/\,3.62$\times$ & 4.75\,/\,3.09$\times$ & 5.42\,/\,3.59$\times$ & 4.61\,/\,2.92$\times$ & 4.64\,/\,2.96$\times$ & 3.76\,/\,2.26$\times$ & 3.38\,/\,2.07$\times$ & 4.69\,/\,3.03$\times$ \\
\;\;$+$ \textbf{DLoop}                   & \textbf{10.01}\,/\,\textbf{4.47}$\times$ & \textbf{8.27}\,/\,\textbf{4.18}$\times$ & \textbf{5.96}\,/\,\textbf{3.29}$\times$ & \textbf{9.61}\,/\,\textbf{4.19}$\times$ & \textbf{5.84}\,/\,\textbf{3.20}$\times$ & \textbf{6.25}\,/\,\textbf{3.20}$\times$ & \textbf{4.78}\,/\,\textbf{2.33}$\times$ & \textbf{3.69}\,/\,\textbf{2.09}$\times$ & \textbf{6.80}\,/\,\textbf{3.37}$\times$ \\
\bottomrule
\end{tabular}}
\end{table}

\end{document}